\documentclass{article}

\PassOptionsToPackage{numbers, compress}{natbib}
\usepackage[preprint]{neurips_data_2022}

\usepackage[utf8]{inputenc} % allow utf-8 input
\usepackage[T1]{fontenc}    % use 8-bit T1 fonts
\usepackage{hyperref}       % hyperlinks
\usepackage{url}            % simple URL typesetting
\usepackage{booktabs}       % professional-quality tables
\usepackage{amsfonts}       % blackboard math symbols
\usepackage{nicefrac}       % compact symbols for 1/2, etc.
\usepackage{microtype}      % microtypography
\usepackage{xcolor}         % colors

\usepackage{transparent}
\usepackage{pifont}
\usepackage{amsthm}
\usepackage{amssymb}
\usepackage{amsmath}
\usepackage{tcolorbox}
\usepackage{soul}
\usepackage[ruled]{algorithm2e}
\usepackage{setspace}
\usepackage{tikz}
\usepackage{enumitem}
\usepackage{subcaption}
\usepackage{wrapfig}
\usepackage{multirow}
\usepackage{adjustbox}
\usepackage{tablefootnote}

\SetKwComment{Comment}{/* }{ */}
\DontPrintSemicolon

\hypersetup{
	colorlinks=true,
	linkcolor=blue,
	filecolor=magenta,      
	urlcolor=cyan,
	citecolor=blue,
	pdfpagemode=FullScreen,
}

\newcommand{\vx}{x}
\newcommand{\rvx}{X}

\newcommand{\ry}{Y}

\newcommand{\sR}{\mathbb{R}}
\newcommand{\sN}{\mathbb{N}}
\newcommand{\sX}{\mathcal{X}}
\newcommand{\sY}{\mathcal{Y}}
\newcommand{\sD}{\mathcal{D}}
\newcommand{\evx}{x}
\newcommand{\rw}{W}
\newcommand{\ete}{\hat{\tau}}
\newcommand{\epo}{\hat{\mu}}
\newcommand{\Expect}[1]{\mathbb{E}_{\rvx \sim P} \left[ #1 \right]}
\newcommand{\cmark}{\ding{51}}%
\newcommand{\xmark}{\transparent{0.2}\ding{55}}%
\newcommand{\abs}[1]{\left| #1 \right|}
\newcommand{\prog}{\mu_{\mathrm{prog}}}
\newcommand{\pred}[1]{\mu_{\mathrm{pred#1}}}
\newcommand{\I}{\mathcal{I}}
\newcommand{\Iprog}{\I_{\mathrm{prog}}}
\newcommand{\Ipred}{\I_{\mathrm{pred}}}
\newcommand{\predscale}{\omega_{\mathrm{pred}}}
\newcommand{\gaussian}{\mathcal{N}}
\newcommand{\Attrpred}{\mathrm{Attr}_{\mathrm{pred}}}
\newcommand{\Attrprog}{\mathrm{Attr}_{\mathrm{prog}}}
\newcommand{\Dtrain}{\sD_{\mathrm{train}}}
\newcommand{\Dtest}{\sD_{\mathrm{test}}}

\newcommand*\circled[1]{\tikz[baseline=(char.base)]{
		\node[shape=circle,draw,inner sep=.3pt] (char) {#1};}}
\newcommand{\alphaprog}{\alpha_{\mathrm{prog}}}	
\newcommand{\alphapred}[1]{\alpha_{\mathrm{pred#1}}}	
\newcommand{\nonlinscale}{\omega_{\mathrm{nl}}}

\theoremstyle{definition}

\newcommand{\squish}[1]{{#1\parfillskip=0pt\par}}

\title{Benchmarking Heterogeneous Treatment Effect Models through the Lens of Interpretability}

\author{%
 Jonathan Crabbé\thanks{Equal contribution. Contact: jc2133@cam.ac.uk, amc253@cam.ac.uk,  ioana.bica@eng.ox.ac.uk.} \\
University of Cambridge\\
  \And
  Alicia Curth$^*$ \\
  University of Cambridge \\
  \And 
  Ioana Bica$^*$ \\
  University of Oxford\\
  \And
      Mihaela van der Schaar \\
  University of Cambridge, UCLA
}

\begin{document}

\maketitle

\vspace{-3mm}
\begin{abstract}
  Estimating personalized effects of treatments is a complex, yet pervasive problem. To tackle it, recent developments in the machine learning (ML) literature on heterogeneous treatment effect estimation gave rise to many sophisticated, but opaque, tools: due to their flexibility, modularity and ability to learn constrained representations, neural networks in particular have become central to this literature. Unfortunately, the assets of such black boxes come at a cost: models typically involve countless nontrivial operations, making it difficult to \textit{understand} what they have learned. Yet, understanding these models can be crucial -- in a medical context, for example, discovered knowledge on treatment effect heterogeneity could inform treatment prescription in clinical practice. In this work, we therefore use post-hoc \textit{feature importance} methods to identify features that influence the model's predictions. This allows us to evaluate treatment effect estimators along a new and important dimension that has been overlooked in previous work: We construct a benchmarking environment to empirically investigate the ability of personalized treatment effect models to identify \textit{predictive covariates} -- covariates that determine differential responses to treatment. Our benchmarking environment then enables us to provide new insight into the strengths and weaknesses of different types of treatment effects models as we modulate different challenges specific to treatment effect estimation -- e.g. the ratio of prognostic to predictive information, the possible nonlinearity of potential outcomes and the presence and type of confounding.

\end{abstract}

\section{Introduction} \label{sec:intro}
The need to estimate the effects of actions -- such as treatments, policies and other interventions -- is ubiquitous in many domains, ranging from economics to medicine. Many applications where treatment effects are of interest additionally operate under \textit{high stakes}, for example treatment decisions in a hospital setting -- making it particularly important that estimates leading to individual decisions are reliable.  %\hl{ADD: something about high stakes applications.} 
As interest in designing \textit{personalized} treatments is growing across fields, a substantial literature on learning treatment effect heterogeneity has emerged in machine learning (ML) in recent years. In this context, a plethora of sophisticated methods for estimating \emph{conditional average treatment effects} (CATE) and/or an individuals \emph{potential outcomes} (POs) under different treatments have been proposed in the recent ML literature (see e.g. \cite{bica2021real}), all aiming to improve the \textit{precision} in estimating effects. In particular, recent work has produced both \textit{model-agnostic} estimation strategies, which can incorporate \textit{any} ML method into the estimation of effects \cite{kunzel2019metalearners, Kennedy2020OptimalDR, curth2021nonparametric}, \textit{and} model-specific strategies, which \textit{adapt specific} ML methods to the treatment effect estimation problem -- for example random forests \cite{wager2018estimation, athey2019generalized}, gaussian processes \cite{alaa2017bayesian, alaa2018limits} and, most predominantly, neural networks \cite{curth2021nonparametric, johansson2016learning, shalit2017estimating, hassanpour2019learning,  assaad2021counterfactual, curth2021inductive}. 

%\hl{Many learning strategies + ref}

\squish{Despite the growing interest in such methods, \cite{curth2021really} noted that their evaluation has been quite one-dimensional: Most, if not all, of this work has focussed on assessing performance of proposed algorithms in terms of the \emph{Precision of Estimating Heterogeneous Effects} (PEHE) criterion of \cite{hill2011bayesian}, which measures the root mean squared error (RMSE) of the estimated CATE function on a test-set. However, in many applications, black-box predictions of expected treatment effects do not suffice: in the context of drug development, for example, it is at least equally important to assess whether an algorithm \textit{discovers the correct drivers of the underlying effect heterogeneity} or leads to the right \textit{interpretation} thereof \cite{hermansson2021discovering}. Such interpretation could also be important in clinical practice e.g. when \textit{explaining} treatment recommendations derived from an estimated CATE function.  In this paper, we therefore leverage recent advances in \emph{explainable artificial intelligence}~\cite{BarredoArrieta2020, Tjoa2020, Das2020} to interpret the discoveries of ML-based CATE estimators, and propose a benchmarking environment which we use to provide insight into the performance of different learning strategies in discovering drivers of heterogeneity.}% -- a new yet important task to evaluate such estimators on.

\begin{figure}
  %\vspace{-3mm}
  \centering
  \includegraphics[width=0.87\textwidth]{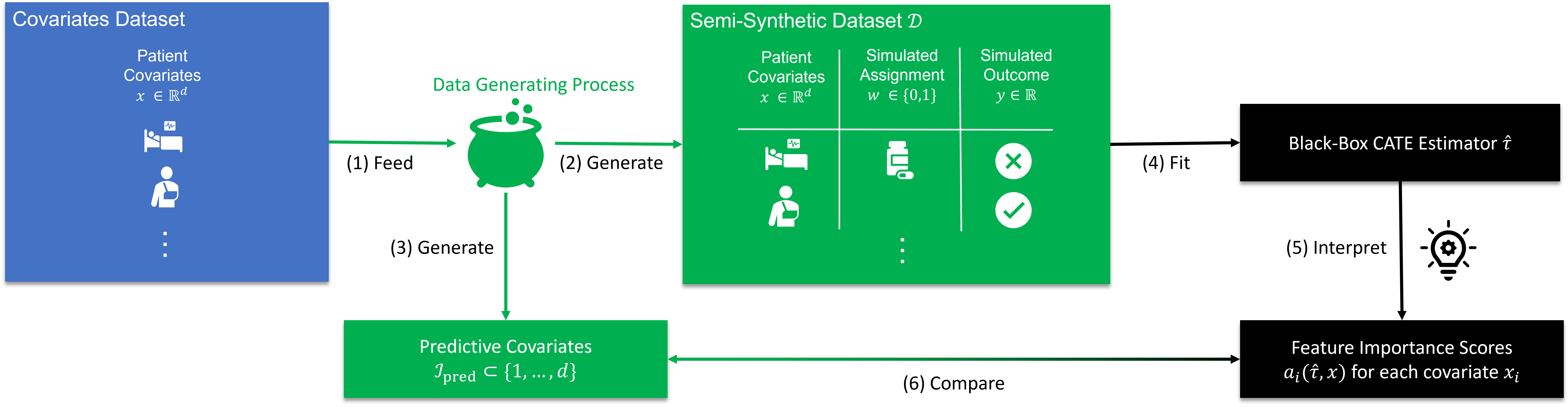}
  \caption{Illustration of the ITErpretability benchmarking environment. (1) Covariates are extracted from any dataset. (2) These covariates are labeled with a transparent data generating process for the assignments and the outcomes. This results in a semi-synthetic dataset $\sD$. (3) Since the data generating process is transparent, we know the indices $\Ipred$ of predictive covariates. (4) We fit a black-box CATE estimator $\ete$ on $\sD$. (5) We then use a feature importance method to assign feature importance scores $a_i(\ete, x)$ to each covariate $x_i$. (6) We evaluate each CATE estimator based on a new metric measuring the accordance between the most important features and the predictive covariates.} 
  \label{fig:benchmark_visual}
  \vspace{-7mm}
\end{figure}

%if the goal is using an algorithm for the design of personalized treatment plans, we would argue that it is at least equally important to assess whether an algorithm leads to the right \textit{interpretation} of the drivers of the underlying effect heterogeneity.

\squish{\textbf{Related work. } How to interpret CATE estimators has received little attention in the ML literature thus far. Some work \textit{implicitly} enables interpretation by relying on methods that are inherently more interpretable, e.g. linear regressions \cite{imai2013estimating, hahn2018regularization} or tree-based models \cite{Foster2011SubgroupIF, athey2016recursive}. Another recent stream of work \textit{explicitly} focuses on interpretability, similarly proposing the use of methods that are inherently interpretable: \cite{lakkaraju2017learning} rely on decision lists, \cite{morucci2020adaptive} construct interpretable hyper-boxes for matching, \cite{lee2020causal} use causal rule ensembles, \cite{nagpal2020interpretable} use mixture models with sparse components, \cite{padilla2021causal} use the fused lasso for estimation of subgroup piece-wise constant treatment effects and \cite{chen2022covariate} rely on an additive neural network architecture. With a goal similar to our work, \cite{hermansson2021discovering} investigate how well different \textit{forest-based} CATE estimation strategies discover effect modifiers, but they rely on variable importance scores inherent to random forests to do so.  To the best of our knowledge, the only work that considers \textit{post-hoc} interpretability of already fitted, arbitrary black-box CATE models uses \textit{model distillation} approaches to do so: \cite{wu2021distilling} use the fitted black-box CATE estimator as a ``teacher'' and a multi-task decision tree as an interpretable ``student'' and similarly, \cite{kim2019learning} propose to create interpretable CATE estimators by fitting an arbitrary interpretable model on top of the potential outcome predictions of a NN-based first stage estimator. Such approaches do not provide an interpretation of the black-box output directly and therefore need to rely on the interpretable student model being good enough to cover the complexities of the original estimator. In this paper, we take a different approach and consider the use of \textit{post-hoc feature importance methods} to interpret black-box CATE estimators \textit{directly}.}

\squish{\textbf{Contributions \& Outlook.} Our contributions are threefold: \circled{1} we study how to \textit{interpret} black-box CATE estimators and use this to \textit{evaluate} them along a \textit{new and important dimension} that has been overlooked in previous work -- namely, their ability to correctly discover drivers of effect heterogeneity --, \circled{2} we propose a \textit{benchmark environment} to do so, and \circled{3} we provide \textit{new insights} into the performance of existing methods on this new task. We proceed as follows: We begin by recalling fundamentals of the CATE estimation setting and discuss its unique characteristics in Section~\ref{sec:cate_estimation}. In Section~\ref{sec:feature_importance}, we review feature importance methods and discuss how they can be applied to interpret what CATE estimators have learned about treatment effect heterogeneity, with the ultimate goal to \textit{compare different CATE estimators} on their ability to identify predictive covariates/features (i.e. covariates that determine differential responses to treatment and, hence, are the ones that truly matter when estimating treatment effects). In Section~\ref{sec:benchmark}, we then introduce a new benchmark environment for evaluating black-box CATE estimators, supplemented with feature importance methods to interpret their output,  on precisely this task.  This \textit{ITErpretability} benchmark, as illustrated in Figure~\ref{fig:benchmark_visual}, can be used with any covariates dataset and is semi-synthetic: it relies on real covariates with synthetic treatment assignment and outcomes, supplying us with otherwise unavailable \textit{ground-truth access} to predictive covariates. Finally, in Section ~\ref{sec:experiments}, we use this benchmark to investigate the performance of existing, neural network-based, CATE estimation strategies on this new task, and provide interesting insights into the impact of typical difficulties in CATE problems: \circled{1}~the strength of predictive relative to prognostic information, \circled{2}~the nonlinearity of potential outcomes and \circled{3}~the presence of confounding.}

\vspace{-2mm}
\section{Setting: The CATE Estimation Problem} \label{sec:cate_estimation}
We consider a standard treatment effect estimation setting as formalized within the Neyman-Rubin potential outcomes framework \cite{rubin2005causal}. We assume access to an observational i.i.d. dataset $\sD=\left\{\left(\ry^n, \rvx^n, \rw^n \right)\right\}_{n=1}^{N}$ %with $\left(\ry^n, \rvx^n, \rw^n \right) \stackrel{i . i . d .}{\sim} P$. 
Here, $\ry^n \in \sY \subset \sR$ is a binary or continuous outcome of interest, $\rvx^n \in  \sX \subset \sR^{d}$ is a vector of pre-treatment covariates, and $\rw^n \in\{0,1\}$ is a binary treatment assigned according to a (usually unknown) propensity score $\pi(\vx)=P(\rw = 1 \mid \rvx = \vx)$. Each individual has multiple potential outcomes (POs) $Y^n (w)$ associated with different treatment values, yet only the outcome associated with the received treatment is observed, i.e. $Y^n=Y^n(W^n)$. To investigate treatment effect heterogeneity, we focus on the conditional average treatment effect (CATE), computed as the expected difference between the two POs for an individual with covariates $\rvx = \vx$:
\begin{align} \label{eq:CATE}
\tau(\vx)=\mathbb{E}_{P}[\ry{(1)}-\ry{(0)} \mid \rvx = \vx] = \mu_1(x)-\mu_0(x)
\end{align}
with $\mu_w(x)=\mathbb{E}_{P}[\ry{(w)} \mid \rvx = \vx]$ the expected potential outcome. 

\textbf{What Makes CATE Estimation Special?} As we discuss in more detail in Appendix~\ref{appendix:cate_details}, three characteristics are generally considered central to the CATE estimation problem (see e.g. \cite{curth2021inductive, curth2021really}): \circled{1}~The need to rely on strong untestable assumptions to ensure \textit{identifiability} of treatment effects from observational data (here: ignorability assumptions \cite{rosenbaum1983central}). \circled{2}~The presence of \textit{covariate shift} due to confounding (correlation between covariates and treatment assignment). \circled{3}~The \textit{absence of the target label} of interest ($\ry{(1)}-\ry{(0)}$) as only one of the POs can be observed in practice, or conversely, that CATE can arise either as a single regression or as a \textit{difference} between two functions

In our context, \circled{1}~\&~\circled{2} are mainly important for the \textit{construction of benchmarks}: as discussed in Section \ref{sec:benchmark}, characteristic~\circled{1} (and to some extent \circled{3}) leads to the necessity to rely on (semi-)\textit{synthetic} data as ground truth knowledge of the data-generating process is generally unavailable from real data (see also the discussion in \cite{curth2021really}). Characteristic~\circled{2} leads to a natural \textit{experimental knob} to include in a benchmark to be modulated when evaluating models. As we discuss in detail in Section \ref{sec:feature_importance}, it is \circled{3} that leads to most interesting considerations when studying \textit{how to interpret} CATE estimators. It also lead to the emergence of different CATE estimation strategies: generally, as discussed in \cite{curth2021nonparametric}, $\tau(x)$ can either be estimated \textit{indirectly} using the difference between the PO regression estimates (i.e. as $\hat{\mu}_1-\hat{\mu}_0$) as in most methods proposed in the recent ML literature \cite{alaa2017bayesian, alaa2018limits, shalit2017estimating, assaad2021counterfactual, curth2021inductive} or \textit{directly} by fitting a regression model using pseudo-outcome $\ry_{\hat{\eta}}$ as a surrogate for the unobserved PO difference (relying on initial estimates of some of the nuisance functions $\eta =(\mu_0, \mu_1, \pi)$) as in e.g. \cite{kunzel2019metalearners, Kennedy2020OptimalDR, curth2021nonparametric}. We discuss different instantiations of these strategies considered in our experiments in Section \ref{sec:experiments}.

\vspace{-2mm}
\section{Feature Importance in the Context of Treatment Effect Estimation} \label{sec:feature_importance}
In this section, we discuss how to use feature importance to investigate what a CATE estimator (or: model) has learned about treatment effect heterogeneity. We start by describing a general feature importance formalism: We assume that we have access to a \textit{black-box} model $\ete: \sR^d \rightarrow \sR$ to estimate the CATE $\tau$. Feature importance methods permit to understand the prediction of this model by highlighting features (covariates) the model is sensitive to. Concretely, this is done by assigning an importance score $a_i (\ete, x) \in \sR$ to each feature $x_i$  contained in the vector $x$ with~\footnote{In the following, $[n]$ denotes the set of natural numbers between $1$ and $n \in \sN^*$.} $i \in [d]$. This score reflects the importance of $x_i$ to predict the CATE $\ete(x)$. The score is such that the importance of a feature $x_i$ increases with $\abs{a_i (\ete, x)}$. When it comes to the sign, features with $a_i (\ete, x) > 0$ tend to increase the CATE $\ete(x)$ and features with $a_i (\ete, x) < 0$ tend to decrease it. 

There exist many methods to assign importance scores $a_i (\ete, x)$ to the features. Different feature importance methods tend to attribute different relative importance between features \cite{Yang2019, Saarela2021}. This is because they measure different characteristics of the black-box model: \emph{gradient-based} methods compute scores based on the model's gradient with respect to the features~\cite{Simonyan2013, Sundararajan2017, Lundberg2017, Erion2021}; \emph{perturbation-based} methods compute scores based on the model's sensitivity to features perturbations~\cite{Ribeiro2016, Fong2017, Crabbe2021Dynamask} and some other methods rely on the neuron's activation to compute the importance scores~\cite{Binder2016, Shrikumar2017}. 
In Section~\ref{sec:experiments}, we use Integrated Gradients~\cite{Sundararajan2017} as our main feature importance method. This is because it offers the best performances empirically and is typically more computationally efficient than the previous methods. A comparison between different feature importance methods is provided in Appendix~\ref{appendix:feature_importance}. We will now discuss the specificity of CATE models in an interpretability perspective.

%In our experiments, we will use Integrated Gradients~\cite{Sundararajan2017} that are endowed with interesting properties for CATE models. We elaborate the rationale behind this choice and compare different feature importance methods in Appendix~\ref{appendix:feature_importance}. 

\squish{\textbf{What Makes CATE Interpretability Special?} Above, we fixed the black-box to interpret to be the CATE estimate $\ete$. However, CATE can of course also be written (and, as in most methods proposed in the recent ML literature, be estimated) in terms of the expected POs, i.e. $\tau = \mu_1 - \mu_0$, which have their own feature importance $a_i (\mu_0, x)$ and $a_i (\mu_1, x)$. In general, feature importance for CATE will differ from those of the POs, leading to different insights. This multiplicity of interpretation certainly distinguishes CATE models from interpreting standard supervised learners and deserves a discussion.}

In order to attach meaning to those possible interpretations, it is useful to consider an important distinction between so-called \textit{predictive} and \textit{prognostic} covariates made in the medical literature~\cite{ballman2015biomarker, sechidis2018distinguishing}. Prognostic covariates determine outcome regardless of treatment assignment -- common risk factors such as age or gender may fall in this category. Such variables are taken into account by the two potential outcomes $\mu_{w}$ in a similar way. In this way, prognostic covariates correspond to features $x_i$ that are important for both POs $a_i (\mu_0, x) \neq 0$ and $a_i (\mu_1, x) \neq 0$. Predictive covariates, on the other end, are predictive of effect heterogeneity, i.e they determine differential responses to treatment -- hormone receptor status in cancer patients is one such example \cite{rastelli2008factors}. In this way, predictive covariates correspond to features $x_i$ that are important for the CATE $a_i (\tau, x) \neq 0$. In the medical context, any measured patient covariate could be prognostic, predictive or both (or, of course, irrelevant).

\squish{Whereas \cite{kim2019learning} used a model-distillation approach to interpret only the PO models,  we argue that it is most interesting in the CATE estimation context to interpret the learned CATE models \textit{directly} by focusing on the discovery of predictive covariates. Identifying such predictive covariates can, for example, provide precious information to support exploratory analyses in clinical trials, which in turn allow pharmaceutical companies to refine the target population for a treatment, improving the likelihood of a successful later stage trial \cite{hermansson2021discovering}. From that perspective, CATE models $\ete$ that are better at identifying predictive covariates are clearly preferable. Due to the absence of treatment effect labels (Section~\ref{sec:cate_estimation}), it is far from obvious whether all estimation strategies result in successful identification of predictive covariates. Furthermore, since indirect learners target the CATE indirectly through the POs, there is no guarantee that the resulting models can distinguish between prognostic and predictive covariates. With these simple observations, it is obvious that CATE interpretability comes with a unique set of challenges. We will now introduce a benchmark to study CATE models through this angle.}

\vspace{-2mm}
\section{The ITErpretability Benchmark} \label{sec:benchmark}
\squish{Next, we describe our proposed \textit{ITErpretability} benchmark that uses ideas from \textit{interpretability} to measure the ability of \textit{treatment effect} estimators to identify predictive covariates. We propose a framework that relies on  a \textit{semi-synthetic} data generating process (DGP), which is standard in the CATE estimation literature \cite{curth2021really}: because identifiability assumptions are generally untestable, \textit{simulating} outcomes and treatment assignments ensures that they hold; additionally it ensures that the underlying CATE function is \textit{known}. In our context, we cannot rely on existing and established semi-synthetic benchmarks such as IHDP \cite{hill2011bayesian, shalit2017estimating} or ACIC2016 \cite{dorie2019automated}, most importantly because they did not \textit{record} which covariates are predictive or prognostic. Further, the \textit{experimental knobs} considered therein are not of primary interest in our setting\footnote{In fact, due to the DGP used in IHDP \textit{all} important variables are both predictive and prognostic, allowing for no interesting distinctions between discoveries.}; instead we thus design our own DGPs that allow to us to obtain interesting new insights in our experiments. Below, we discuss our DGP and proposed metrics.}

\squish{\textbf{DGP.} We would like to rely on a DGP that covers a range of realistic scenarios \textit{and} for which we can clearly identify prognostic and predictive covariates. Since this last information is generally not available in \textit{real} observational data, we use a semi-synthetic approach in which we reuse covariates $X^n$ from a real dataset and synthetically generate the treatment assignments $W^n$ and outcomes $Y^n$ for all $n \in [N]$. In this way, the resulting semi-synthetic dataset has realistic covariates \textit{and} we have a full knowledge on how outcomes are generated. In particular, we can restrict to DGPs for which prognostic and predictive covariates are clearly distinct and identifiable. We implement this by selecting \textit{non-overlapping} subsets $\Iprog \subset [d]$ of prognostic and two subsets $\I_0 , \I_1 \subset [d]$ of predictive covariates. The prognostic covariates similarly contribute to both POs $Y{(0)}, Y{(1)}$ through a function $x \mapsto \prog(x_{\Iprog})$, where we let $x_{\I}$ denote the vector $(x_i)_{i \in \I}.$ for a set $\I \subset [d]$. The predictive covariates, on the other hand, contribute to either only $Y{(0)}$ through a function $x \mapsto \pred{0}(x_{\I_{0}})$ or to $Y{(1)}$ through a function $x \mapsto \pred{1}(x_{\I_{1}})$.\footnote{Note that the expected difference between the POs can be written as $\mathbb{E}[Y{(1)} - Y{(0)} \mid X = x] = \predscale \cdot [\pred{1}(x_{\I_1}) - \pred{0}(x_{\I_0})]$. In this way, the treatment effect depends only on the covariates indexed by $\Ipred = \I_0 \sqcup \I_1$. This indeed corresponds to the definition of predictive covariates given in Section~\ref{sec:feature_importance}.} We include a predictive scale $\predscale \in \sR^+$ that permits to tune the relative strength between the prognostic and predictive contributions to the POs. This full process is detailed in Algorithm~\ref{alg:dgp}. All the experiments from Section~\ref{sec:experiments} are produced by varying the inputs of this algorithm; in particular, we consider different types of outcome functions and propensity scores.}
    \vspace{-2mm}
\begin{algorithm}
	\setstretch{1}
	\caption{Semi-Synthetic Data Generating Process}\label{alg:dgp}
	\KwIn{Covariates dataset $\{ X^n \in \sR^d \}_{n=1}^N $, Prognostic function $\prog : \sR \rightarrow \sR$, Predictive functions $\pred{0}, \pred{1} : \sR \rightarrow \sR$, Propensity score $\pi : \sX \rightarrow \sR$, Feature sets size $n_{\I} \in \sN^*$, Predictive scale $\predscale \in \sR^+$, Noise level $\sigma \in \sR^+$} 
	\KwOut{Semi-synthetic observational dataset $\sD = \{ (Y^n , X^n , W^n) \}_{n=1}^N$, Prognostic features $\Iprog \subset [d]$, Predictive features $\Ipred \subset [d]$}
	\textbf{Ensure:} $d > 3 \cdot n_{\I}$ \Comment*[r]{Avoid overlap between $\Iprog, \ \I_0$ and $\I_1$}
	$\I \ \gets$ Sample $3\cdot n_{\I}$ elements from $[d]$ without replacement \Comment*[r]{Get relevant features}
	$\Iprog, \I_0 , \I_1 \gets$ Split $\I$ into 3 sets of size $n_{\I}$\Comment*[r]{Get prog. and pred. features}
	$\sD \gets \emptyset$\Comment*[r]{Initialize dataset}
	\For{$n \in [N]$}{
	$Y{(0)} \gets \prog(X^n_{\Iprog}) + \predscale \cdot \pred{0}(X^n_{\I_0})$\Comment*[r]{Get untreated outcome}
	$Y{(1)} \gets \prog(X^n_{\Iprog}) + \predscale \cdot \pred{1}(X^n_{\I_1})$\Comment*[r]{Get treated outcome}
	$W^n \sim \mathrm{Bernoulli}[\pi(X^n)]$\Comment*[r]{Sample treatment assignment}
	$\epsilon \sim \gaussian(0, \sigma)$\Comment*[r]{Sample noise}
	$Y^n = W^n \cdot Y{(1)} + (1 - W^n) \cdot Y{(0)} + \epsilon$\Comment*[r]{Get observed outcome}
	$\sD \gets \sD \cup \{ (Y^n , X^n , W^n) \}$\Comment*[r]{Append dataset}
	}
	\KwRet{$\sD , \ \Iprog , \ \Ipred = \I_0 \sqcup \I_1 $}
\end{algorithm}
\vspace{-2mm}

\textbf{Metrics.} After using Algorithm \ref{alg:dgp}, we split the generated dataset into a training set $\Dtrain$ and a testing set $\Dtest$ ($80 \% - 20 \%$ split). We fit a model $\ete$ to estimate the CATE on the training set $\Dtrain$ and evaluate the model on the held-out test set $\Dtest$. As aforementioned, our purpose is to assess if this model has correctly identified the predictive covariates. With our choice of DGP, we know that those predictive covariates correspond to the indices in $\Ipred= \I_0 \sqcup \I_1$. By recalling that the importance attributed to covariate $x_i$ for a prediction $\ete(x)$ increases with the absolute value  $\abs{a_i(\ete, x)}$, we can compute the average proportion of the attribution \textit{correctly} allocated to the predictive covariates:
\begin{align} \label{eq:attr_pred}
    \Attrpred = \frac{1}{\abs{\Dtest}} \sum_{X \in \Dtest} \frac{\sum_{i \in \Ipred} \abs{a_i(\ete, X)}}{\sum_{i=1}^d \abs{a_i(\ete, X)}}.
\end{align}
Note that $\Attrpred \in [0,1]$, where $\Attrpred = 0$ corresponds to a model that does not identify any predictive covariate and $\Attrpred = 1$ corresponds to a model that only identifies predictive covariates. Ideally, predictive covariates should be the most important for a model that estimates the CATE; thus we expect good models to score high with respect to this metric. A similar metric $\Attrprog$ can be defined analogously to measure the fraction of the feature attribution \textit{incorrectly} allocated to the prognostic covariates by replacing $\Ipred \mapsto \Iprog$ in~\eqref{eq:attr_pred}. Ideally, $\Attrprog$ should be zero if all importance is correctly allocated to predictive variables. Finally, we will sometimes also report the standard PEHE metric, i.e. the RMSE of estimating CATE: $\mathrm{PEHE}=\sqrt{n_{\text{test}}^{-1}\sum_{X \in \Dtest} [\hat{\tau}(X)-\tau(X)]^2}$.

\iffalse
\begin{align}
    \Attrprog = \frac{1}{\abs{\Dtest}} \sum_{X \in \Dtest} \frac{\sum_{i \in \Iprog} \abs{a_i(\ete, X)}}{\sum_{i=1}^d \abs{a_i(\ete, X)}}.
\end{align}
\fi

\vspace{-2mm}
\section{Experiments} \label{sec:experiments}
\vspace{-2mm}
In this section, we benchmark different types of CATE estimators on their ability to identify predictive covariates through feature importance scores. We study 3 different characteristics of the data generating process which we expect to impact this ability: \circled{1} The relative strength between the prognostic contribution $\prog$ and the predictive contributions $\pred{0}, \pred{1}$ to the POs $Y{(0)}, Y{(1)}$ (Sec.~\ref{sec:exp_pred_scl}); \circled{2} The presence of nonlinearities in the prognostic and predictive functions $\prog, \pred{0}, \pred{1}$ (Sec.~\ref{sec:exp_nonlin}), and \circled{3} the fact that the treatment assignment might be biased according to a nontrivial propensity score $\pi$ (Sec.~\ref{sec:exp_conf}). All the experiments are done by varying the inputs of Algorithm~\ref{alg:dgp}.

\textbf{Datasets.} We extract covariates to use in our our benchmarking environment from the following four datasets: TCGA \cite{weinstein2013cancer}, Twins \cite{almond2005costs}, News \cite{newman2008bag} and ACIC2016 \cite{dorie2019automated}, which were selected due to their diverse characteristics in terms of the number of features, mixture of categorical/continuous features and population size. The number of covariates in these datasets range from $d=39$ to $d=100$ and we set $n_{\mathcal{I}} = \lfloor 0.2 \cdot d \rfloor$. Refer to Appendix \ref{appendix:experiments_details} for details of the datasets. 

%\hl{TBD: @Ioana can you add this description?  maybe also add quick comment on how these differ; in terms of feature types, sample sizes etc. any reasons why we chose specifically these datasets? We should also add references for the paper checklist.}

\squish{\textbf{Learners.} We consider a number of CATE estimators based on neural networks throughout our experiments. As \textit{direct} estimators, we use \cite{Kennedy2020OptimalDR}'s DR-learner, which relies on a doubly robust pseudo-outcome, and \cite{kunzel2019metalearners}'s X-learner, which uses a weighted average of two direct (singly-robust) treatment effect estimates from both treatment groups. As \textit{indirect} estimators, we use \cite{kunzel2019metalearners}'s T-learner, which fits \textit{t}wo regression models $\hat{\mu}_w(x)$ (one for each treatment group) and sets $\hat{\tau}(x)=\hat{\mu}_1(x)-\hat{\mu}_0(x)$, and S-learner, which includes the treatment assignment variable $W$ as a standard feature in a \textit{s}ingle regression $\hat{\mu}(x, w)$ and sets $\hat{\tau}(x)=\hat{\mu}(x, 1) - \hat{\mu}(x, 0)$. Finally, we consider \cite{shalit2017estimating}'s TARNet which can be seen as a \textit{hybrid} between S- and T-learner \cite{curth2021nonparametric}: it learns a  representation $\Phi(x)$ \textit{shared} between treatment groups, which is used by \textit{treatment-specific} outcome heads $h_w(\Phi(x))$ so that $\hat{\tau}(x)=h_1(\Phi(x)) - h_0(\Phi(x))$. In the experiments with confounding, we also use \cite{shalit2017estimating}'s CFRNet, which differs from TARNet only in a regularization term that encourages the representation to be \textit{balanced} (follow a similar distribution) across treatment groups.  We discuss all models in more detail in Appendix \ref{appendix:experiments_details}, where we also detail how we fix hyperparameters across all models to ensure similar capacity.}

\vspace{-2mm}
\subsection{Experiment 1: Altering the Strength of Predictive Effects}\label{sec:exp_pred_scl}

\textbf{Setup.} We begin by investigating how the strength of predictive effects \textit{relative} to prognostic effects influences the ability of different learners to discover predictive covariates. This is an interesting question, as in practice predictive signals are often assumed to be much \textit{weaker} than prognostic ones \cite{kunzel2019metalearners, curth2021inductive, hermansson2021discovering}.  Thus, ideally estimators should be able to correctly identify predictive covariates even when effects are weak; yet some of \cite{hermansson2021discovering}'s empirical results comparing different random forest-based learners across DGPs with different predictive effects, show that this is not always the case. For our experiments, we thus continuously vary predictive effect size in a DGP with a linear parametrization for the prognostic and predictive functions: $\prog(x_{\Iprog}) = \alphaprog^{\intercal} \ x_{\Iprog} , \ \pred{0}(x_{\I_0}) = \alphapred{0}^{\intercal}  \ x_{\I_0} $ and $\pred{1}(x_{\I_1}) = \alphapred{1}^{\intercal} \ x_{\I_1}$ with weights sampled randomly $\alphaprog, \ \alphapred{0}, \ \alphapred{1} \sim U([-1, 1]^{n_{\I}})$, where $U$ denotes the uniform distribution. To vary the relative strength between the prognostic and the predictive contributions to the POs, we change the predictive scale $\predscale \in \{10^{-3}, 10^{-2}, 10^{-1}, 0.5, 1\}$. Here, treatments are assigned completely at random, i.e. $\pi(x)= 0.5$.

\begin{figure}[t]
    %\vspace{-4mm}
    \centering
    \includegraphics[width=0.9\textwidth]{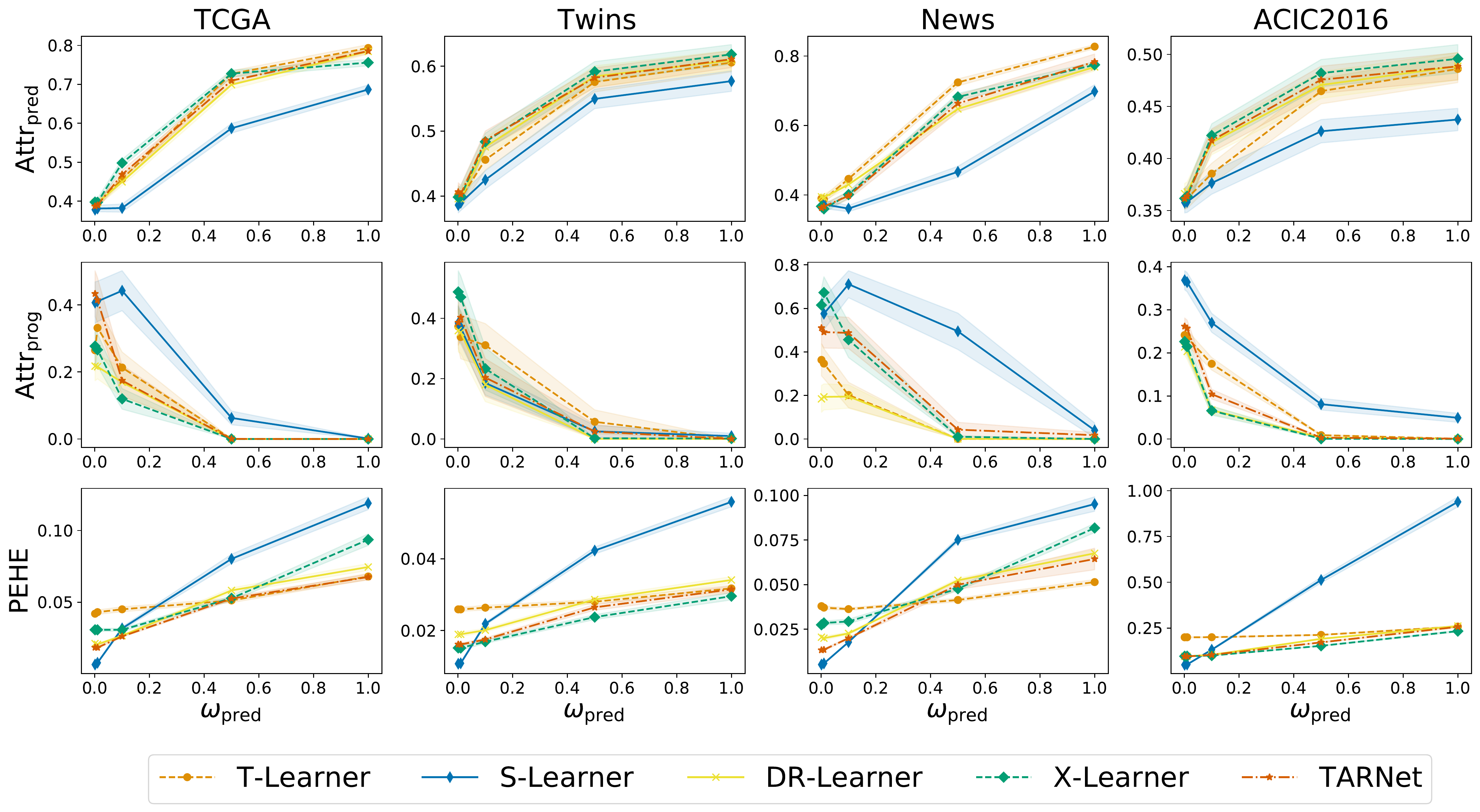}
    \caption{Performance comparison in terms of $\Attrpred$ (top), $\Attrprog$ (middle) and PEHE (bottom) when varying the predictive scale, using four feature datasets (TGCA, Twins, News, ACIC). Averaged across multiple runs, shaded areas indicates one standard error.}
    \label{fig:performance_predictive_sc}
    \vspace{-6mm}
\end{figure}

\squish{\textbf{Results.} In Fig. \ref{fig:performance_predictive_sc} we present results on correct attributions $\Attrpred$ and misattributions $\Attrprog$, as well as the standard PEHE metric for comparison, for all datasets. We make a number of interesting observations: \circled{1}  \textit{Attribution trends indeed vary with $\predscale$.} Correct predictive attributions $\Attrpred$ substantially increase as $\predscale$ increases for all learners and across all datasets considered -- this confirms the intuition that the stronger the predictive effects are, the easier it is to identify their origin. Conversely, prognostic misattributions $\Attrprog$ decrease as $\predscale$ increases, indicating that one reason for the low $\Attrprog$ at low $\predscale$ is that learners \textit{confuse} prognostic effects for predictive ones. \circled{2} \textit{Comparing learners, S-Learners appear to struggle most to make correct attributions}. The most salient observation across all datasets is that the S-Learner does does substantially worse at $\Attrpred$. With the exception of the Twins dataset, this usually also translates into higher $\Attrprog$ than all other learners. We believe that this is because the S-Learner uniquely neither has a treatment-group specific component (like T-learner and TARNet do) nor models CATE directly (like DR- and X-learner do); learned treatment effect heterogeneity thus has to arise through learned interactions with the treatment indicator -- which appears to lead to less reliable predictive covariate discoveries. All other methods perform very similar to each other. \circled{3} \textit{Using attribution metrics indeed leads to interesting new insights relative to considering only PEHE.} We observe that the S-learner does \textit{best} in terms of PEHE when predictive strength is low, while the T-learner does worst (as discussed below, this is in line with expectations %\footnote{Because of their structure, S-Learners learn CATE functions that are \textit{less} heterogeneous/relatively simple while T-learners, because of the lack of shared component, tend to learn very heterogeneous/complex functions even when no heterogeneity exists \cite{kunzel2019metalearners}} 
and empirical observations previously made in e.g. \cite{kunzel2019metalearners, curth2021inductive}) -- yet, as we saw above, this does not translate into better discoveries. Similarly, the better performance in terms of PEHE of some other strategies relative to T-learner when $\predscale$ is small also does not lead to better discoveries. We attribute this to the fact that when $\predscale$ is very small, PEHE will favour any method that outputs near-zero treatment effects; indeed all considered methods except the T-learner incorporate an implicit inductive bias that shrinks effects \cite{curth2021inductive} -- which appears to help only  in terms of PEHE. Note also that PEHE is not directly comparable across different values of $\predscale$ as it naturally increases as the scale of CATE changes.}

\vspace{-2mm}
\subsection{Experiment 2: Incorporating Nonlinearities}\label{sec:exp_nonlin}

\textbf{Description.} In practice, there is no particular reason to expect that POs are linear functions of the covariates. Next we therefore investigate how nonlinearities in the POs influence the ability of CATE estimators to identify predictive covariates. %Ideally, they should be able to correctly identify predictive covariates in the presence of nonlinearities. 
To do so, we use a parametrization for the prognostic and predictive functions that allows us to control the strength of the nonlinearities through parameter $\nonlinscale$: $\prog(x_{\Iprog}) = (1 - \nonlinscale) \cdot \alphaprog^{\intercal} \ x_{\Iprog} + \nonlinscale \cdot \chi(\alphaprog^{\intercal} \ x_{\Iprog}), \ \pred{0}(x_{\I_0}) = (1 - \nonlinscale) \cdot \alphapred{0}^{\intercal} \ x_{\I_0} + \nonlinscale \cdot \chi(\alphapred{0}^{\intercal} \ x_{\I_0})$ and $\pred{1}(x_{\I_1}) = (1 - \nonlinscale) \cdot \alphapred{1}^{\intercal} \ x_{\I_1} + \nonlinscale \cdot \chi(\alphapred{1}^{\intercal} \ x_{\I_1})$ with weights and nonlinearity sampled randomly $\alphaprog, \ \alphapred{0}, \ \alphapred{1} \sim U([-1, 1]^{n_{\I}}), \hspace{5mm} \chi \sim U(\mathcal{F})$, where $\mathcal{F}$ is a set of 10 nonlinear functions $\sR \rightarrow \sR$ (specified in Appendix~\ref{appendix:experiments_details}). To vary the strength of the nonlinearities, we let the nonlinearity scale vary $\nonlinscale \in [0,1]$. We note that $\nonlinscale = 0$ corresponds to linear function as in the previous experiment. On the other hand, $\nonlinscale = 1$ corresponds to purely nonlinear functions. Here, we set $\predscale = 1$ -- a setting for which all learners performed well in the previous experiment and $\pi(x)=0.5$. 

\begin{figure}[t]
 %   \vspace{-2mm}
    \centering
    \includegraphics[width=0.9\textwidth]{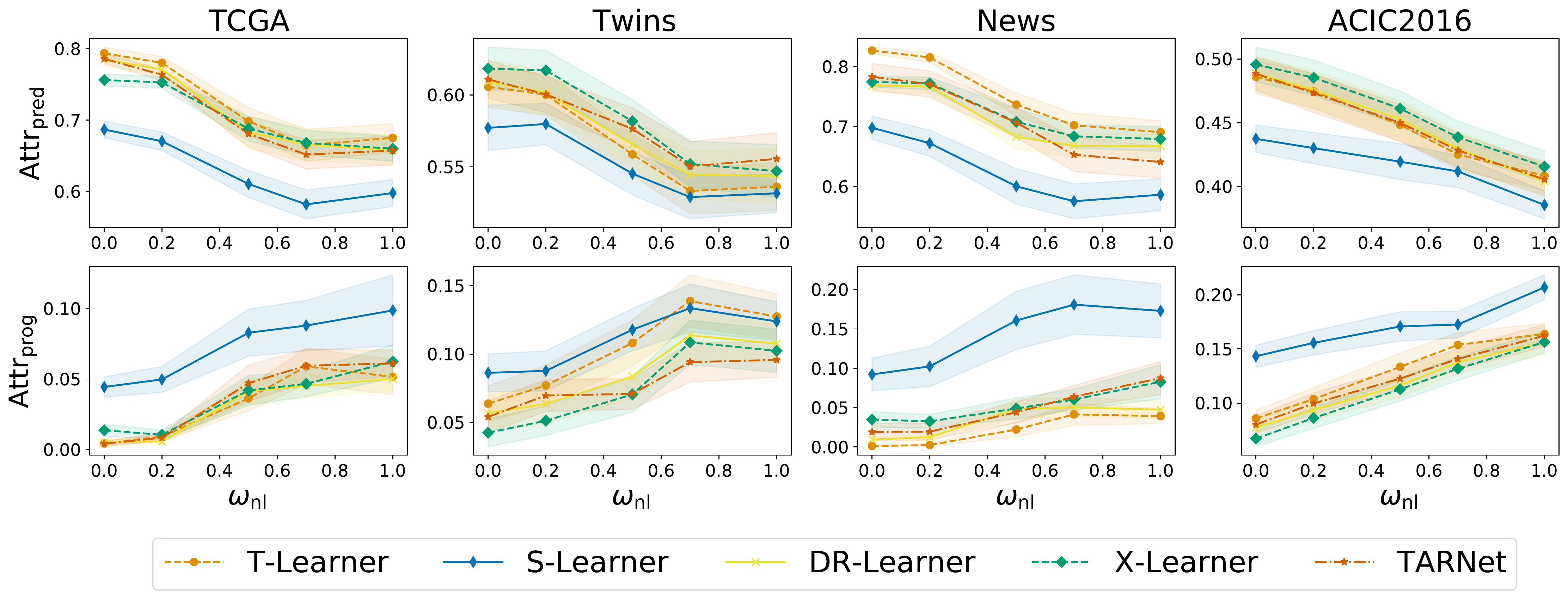}
    \caption{Performance comparison in terms of $\Attrpred$ (top) and $\Attrprog$ (bottom) when varying the nonlinearity scale, using four feature datasets (TGCA, Twins, News, ACIC). Averaged across multiple runs, shaded areas indicates one standard error.}
    \vspace{-7mm}
    \label{fig:performance_nonlinear_sensitivity}
\end{figure}

\textbf{Results.} We present attribution score results across all datasets in Fig. \ref{fig:performance_nonlinear_sensitivity}. We find that \circled{1}  \textit{Attribution trends vary with $\nonlinscale$.} As $\nonlinscale$ increases and the underlying DGP is dominated by the nonlinearity, hence becoming more difficult to learn, we observe that correct attribution ($\Attrpred$) decreases for all methods. Also here, we observe that this is mirrored by an \textit{increase} in confusion of prognostic effects for predictive ones (as seen in the increasing $\Attrprog$). Further, we note that \circled{2} \textit{Relative ordering of methods does not change.} The S-Learner continues to underperform compared to all other learners and the performance gap does not substantially change across values of $\nonlinscale$.

\vspace{-3mm}
\subsection{Experiment 3: The Effect of Confounding}\label{sec:exp_conf}
\textbf{Description.} Finally, we turn to examine the effect of \textit{confounding}, i.e. covariate shift between treatment groups resulting from treatment assignment being based on observables -- a problem that much of the ML literature proposing CATE estimators has focussed on (e.g. \cite{johansson2016learning, shalit2017estimating, assaad2021counterfactual}). A popular solution to deal with said covariate shift has been to rely on \textit{balancing} regularization that penalizes distributional distance (here: MMD$^2$) between treatment groups in representation space; here we therefore also consider \cite{shalit2017estimating}'s CFRNet which is identical to TARNet but includes a discrepancy-based regularization term controlled by hyperparameter $\gamma$. %-- which is difficult to tune in practice as ground truth treatment effects are unavailable for cross-validation. 
We note that if the covariates that determine treatment assignment are either predictive or prognostic -- i.e. they are true \textit{confounders} -- it is generally not possible to remove \textit{all} covariate shift without removing predictive/prognostic information. We therefore consider a final experiment where we structurally vary not only the degree of assignment bias (through propensity scale $\omega_\pi$) but also what \textit{type of information} assignment is based on. We achieve this by modifying the propensity score: $\pi(x) = \texttt{Sigmoid}(\omega_\pi \cdot Z_{\mathrm{score}} [\psi(x)])$, where $Z_{\mathrm{score}}[\cdot]$ indicates normalization across the generated training dataset (this ensures well-behaved propensity scores centered at $0.5$) and $\psi : \sX \rightarrow \sR$ controls the type of confounding. We consider 3 types of confounding, each corresponding to a different choice for $\psi$. \circled{1}~ \textit{Predictive confounding} corresponds to setting $\psi = \pred{1} - \pred{0}$. It mimics a scenario where treatment assignment is biased towards those with characteristics making them most likely to respond well to it, e.g. a doctor assigning treatment with knowledge of CATE. \circled{2}~\textit{Prognostic confounding} corresponds to setting $\psi = \prog$. It mimics the most classical confounding setting where treatment assignment is biased towards those with characteristics making them more likely to have a good outcome regardless of treatment, e.g. self-selection into a treatment program. \circled{3}~Finally, we consider a \textit{non-confounded propensity}: $\psi(x) = x_i$ for some irrelevant covariate $i \notin \Iprog \sqcup \Ipred$. In this case, the treatment selection is not based on a covariate that affects outcome; note that the distribution of covariates in $\Ipred \sqcup \Iprog$ might still differ across treatment groups if they are correlated with the chosen `irrelevant'. Note that for $\omega_\pi=0$, all settings are identical and reduce to the previous $\pi(x) = 0.5$. For the potential outcomes, we consider the previous setting with $\predscale=1$ and $\nonlinscale=0$. We believe that such an explicit distinction between different confounding types has not yet been investigated in related work. 

\begin{wrapfigure}{r}{0.53\columnwidth}
	\vspace{-0.6cm}
    \includegraphics[width=0.53\columnwidth]{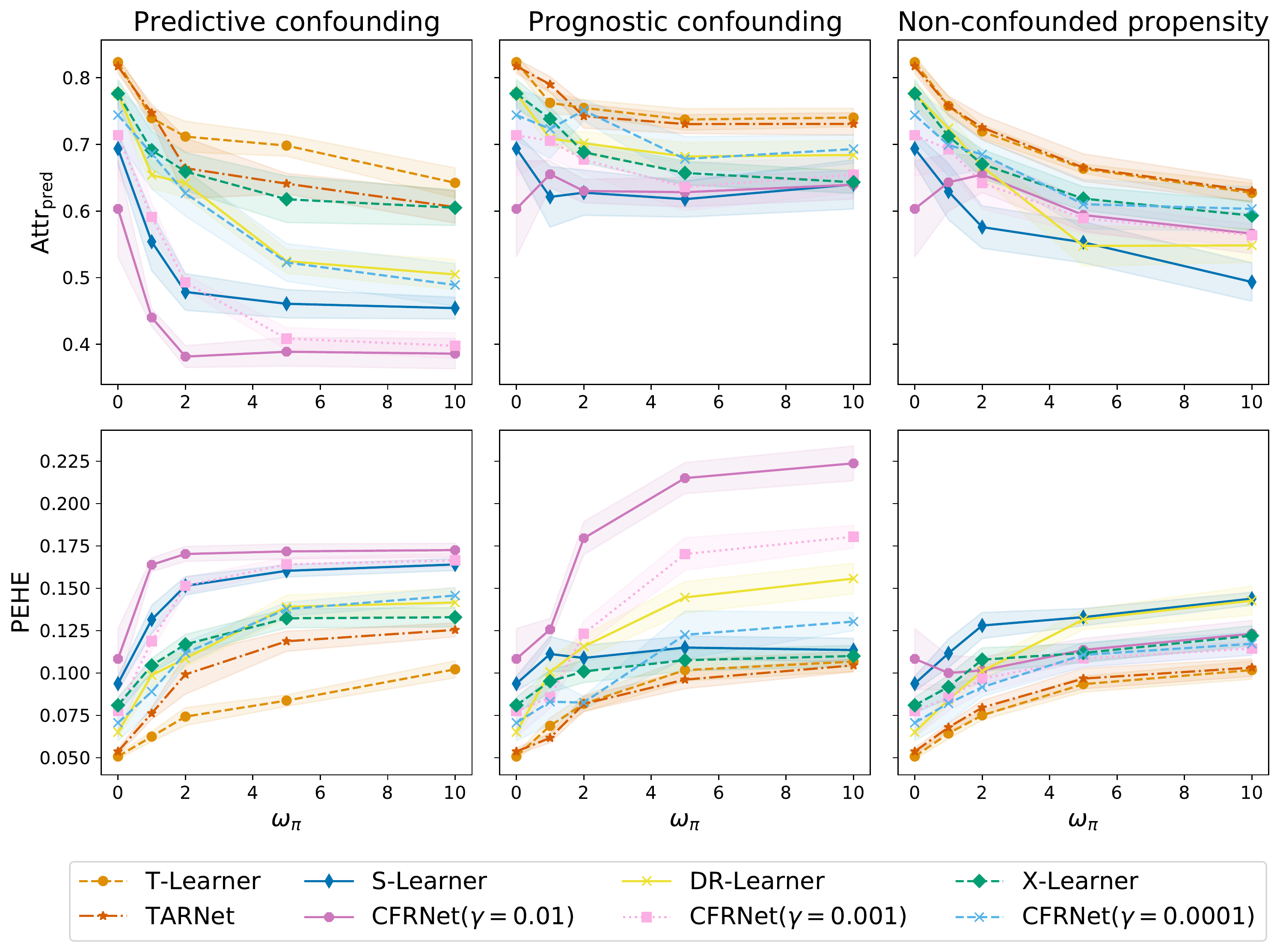}
    \vspace{-0.6cm}
    \caption{Performance comparison when increasing the  propensity scale on the News dataset.}
	\label{fig:performance_prop_sc}
	\vspace{-0.5cm}
\end{wrapfigure}

\textbf{Results.} We present results comparing $\Attrpred$ and PEHE across the three settings for the News dataset in Fig.~\ref{fig:performance_prop_sc} (for full results refer to Appendix~\ref{appendix:experiments_details}). We make numerous interesting observations: \circled{1}~\textit{Attribution trends indeed vary with $\omega_\pi$.} Also in this experiment attribution scores vary systematically as we change $\omega_\pi$; across all propensity scores the quality of attributions worsens the more biased treatment assignment is, which is in line with effects on estimation error.  \circled{2}~\textit{There are systematic differences across the three settings.} We observe that performance deteriorates the most in the setting with predictive confounding. We believe that this is a result of less observed variation across predictive covariates \textit{within} a group due to the assignment bias, making it harder for any model to learn that treatment effect varies systematically across these covariates. Perhaps more surprisingly, we observe that performance as measured by $\Attrpred$ deteriorates  \textit{least} in the prognostic confounding setting (and not in the non-confounded setting). We attribute this to two \textit{competing} forces being at play here: having any sort of covariate imbalance, as in the non-confounded setting, appears to lead to a decrease in $\Attrpred$. In the prognostic confounding setting, there is a similar effect as in the predictive setting that offsets some of that performance decrease: as prognostic covariates now have less variation within a group, models are less likely to \textit{misattribute} a predictive effect to them. \circled{3}~\textit{CFRNet`s balancing regularization has different effects across settings.} We find that the addition of the balancing regularization term can lead to a very large drop in $\Attrpred$ in the predictive confounding setting -- this is expected as aligning distributions should lead to a loss in explanatory power, which is also reflected in the PEHE. In the other two settings, no such effect is visible when considering $\Attrpred$; yet, we observe that PEHE does worsen considerably in the prognostic confounding setting. We believe that this is an effect of the prognostic component of the POs being estimated less accurately due to the balancing regularization, making the CATE estimate, the difference between the two estimated POs, less accurate overall. %\circled{3}\textit{Considering attribution metrics leads to different conclusions than considering only PEHE.} When comparing the three settings in terms of PEHE, we find that the difference between 

\vspace{-2.5mm}
\section{Discussion} \label{sec:discussion}
\vspace{-3.5mm}
\squish{In this paper, we have introduced the ITErpretability benchmark, a new environment to benchmark CATE models with the help of feature importance methods. We empirically demonstrated on various datasets that this benchmark provides insights that are \emph{not} accessible with the metrics and benchmarks considered standard in the CATE literature. We believe that this work opens up many interesting avenues for future research:  First, our environment could be used to extend insights to many more of the CATE estimators proposed in recent literature. One could, for example, replicate our propensity experiments with learners tackling confounding with methods other than balancing, e.g. importance weighting as in \cite{assaad2021counterfactual}, or compare performance across different classes of underlying ML methods -- i.e. compare how discoveries differ across implementations relying on neural networks  with e.g. random forests as in \cite{hermansson2021discovering} or the gaussian processes of \cite{alaa2017bayesian, alaa2018limits}. Second, we believe that it would be interesting to consider how to \textit{improve existing} or \textit{develop new} CATE estimation strategies with the help of interpretability techniques or insights derived from experiments such as the ones from Sec.~\ref{sec:experiments}. Finally, note that we have exclusively focused on \textit{feature importance} methods here. Another possible extension of this work would then be to perform a similar study with other type of explanation methods, such as example-based explanation methods like Influence Functions~\cite{Koh2017} and hybrid methods like SimplEx~\cite{Crabbe2021SimplEx}. }
%Comment from Mihaela:  we can use this benchmarking environment to understand current models and their weaknesses, but also to provide insights which will enable us to build new (better) CATE methods....

\clearpage

\begin{ack}
The authors are grateful to Javier Abad Martinez and Bogdan Cebere for implementing and running preliminary, yet different, experiments in the early and explorative stages of the project. Jonathan Crabbé is funded by Aviva, Alicia Curth is funded by AstraZeneca, Ioana Bica is funded by the Alan Turing Institute (under the
EPSRC grant EP/N510129/1) and Mihaela van der Schaar is funded by the Office of Naval Research (ONR under the grant NSF 1722516). The authors have no competing interests to disclose.  

\end{ack}

\bibliographystyle{unsrt}
\bibliography{neurips_data_2022.bib}
\clearpage

%%%%%%%%%%%%%%%%%%%%%%%%%%%%%%%%%%%%%%%%%%%%%%%%%%%%%%%%%%%%
\section*{Checklist}

\begin{enumerate}

\item For all authors...
\begin{enumerate}
  \item Do the main claims made in the abstract and introduction accurately reflect the paper's contributions and scope?
    \answerYes{All the claims are empirically supported in Section~\ref{sec:experiments}}
  \item Did you describe the limitations of your work?
    \answerYes{Some limitations and possible extensions of our work are discussed in Section~\ref{sec:discussion}.}
  \item Did you discuss any potential negative societal impacts of your work?
    \answerNo{Our benchmark helps to investigate the quality of CATE models, hence mitigating their potential negative impact.}
  \item Have you read the ethics review guidelines and ensured that your paper conforms to them?
    \answerYes{We have reviewed the guidelines and we ensured that our paper is conform.}
\end{enumerate}

\item If you are including theoretical results...
\begin{enumerate}
  \item Did you state the full set of assumptions of all theoretical results?
    \answerNA{Our paper is purely empirical.}
	\item Did you include complete proofs of all theoretical results?
    \answerNA{Our paper is purely empirical.}
\end{enumerate}

\item If you ran experiments (e.g. for benchmarks)...
\begin{enumerate}
  \item Did you include the code, data, and instructions needed to reproduce the main experimental results (either in the supplemental material or as a URL)?
    \answerYes{The code is provided as a supplementary material.}
  \item Did you specify all the training details (e.g., data splits, hyperparameters, how they were chosen)?
    \answerYes{All the training details are provided in Appendix~\ref{appendix:experiments_details} and in the enclosed code.}
	\item Did you report error bars (e.g., with respect to the random seed after running experiments multiple times)?
    \answerYes{All our Figure contain standard errors resulting from running the experiments with different seeds.}
	\item Did you include the total amount of compute and the type of resources used (e.g., type of GPUs, internal cluster, or cloud provider)?
    \answerYes{Our computing ressources are described in Appendix~\ref{appendix:experiments_details}.}
\end{enumerate}

\item If you are using existing assets (e.g., code, data, models) or curating/releasing new assets...
\begin{enumerate}
  \item If your work uses existing assets, did you cite the creators?
    \answerYes{All the relevant references are provided in Section~\ref{sec:experiments}.}
  \item Did you mention the license of the assets?
    \answerYes{The relevant licenses are given in Appendix~\ref{appendix:experiments_details}.}
  \item Did you include any new assets either in the supplemental material or as a URL?
    \answerYes{We provide the implementation of our benchmarking environment in the supplementary materials.}
  \item Did you discuss whether and how consent was obtained from people whose data you're using/curating?
    \answerYes{The relevant discussions are in Appendix~\ref{appendix:experiments_details}.}
  \item Did you discuss whether the data you are using/curating contains personally identifiable information or offensive content?
    \answerNA{All the datasets we use have been de-identified.}
\end{enumerate}

\item If you used crowdsourcing or conducted research with human subjects...
\begin{enumerate}
  \item Did you include the full text of instructions given to participants and screenshots, if applicable?
    \answerNA{No research with human subjects.}
  \item Did you describe any potential participant risks, with links to Institutional Review Board (IRB) approvals, if applicable?
    \answerNA{No research with human subjects.}
  \item Did you include the estimated hourly wage paid to participants and the total amount spent on participant compensation?
    \answerNA{No research with human subjects.}
\end{enumerate}

\end{enumerate}

\clearpage
%%%%%%%%%%%%%%%%%%%%%%%%%%%%%%%%%%%%%%%%%%%%%%%%%%%%%%%%%%%%

\appendix
\section*{Appendix}
This Appendix is structured as follows: in Section \ref{appendix:cate_details}, we give additional background information on CATE estimation. In Section \ref{appendix:feature_importance}, we discuss feature importance methods in more detail. Finally, in Section \ref{appendix:experiments_details} we give further details on experiments and present additional results.

\section{Further Background on CATE Estimation}
\label{appendix:cate_details}
In this section, we discuss the unique characteristics of CATE estimation in more detail. As outlined in section \ref{sec:cate_estimation}, we consider three characteristics most important:

\textbf{\textbullet{ } 1. The need to rely on untestable assumptions. } To infer causal effects from observational data, one needs to make strong \textit{untestable} assumptions  which ensure \textit{identifiability} of a treatment effect and should be assessed by a domain expert in practice. Such assumptions are used to assure that treated and untreated individuals are exchangeable, so that $\mathbb{E}_{P}[\ry{(w)} \mid \rvx = \vx]=\mathbb{E}_{P}[\ry \mid \rw = w , \rvx = \vx]$. As is standard in related literature,  we rely on the strong ignorability conditions \cite{rosenbaum1983central} giving rise to 

\textbf{Assumption 1.} [Consistency, ignorability, and positivity] \emph{Consistency}: If individual $n$ is assigned treatment $w^n$, we observe the associated potential outcome $\ry^{n}=\ry^{n}\left(w^{n}\right)$. \emph{Ignorability}: there are no hidden confounders, such that $\ry{(0)}, \ry{(1)} \perp \rw \mid \rvx$. \emph{Positivity}: treatment assignment is non-deterministic, i.e. $0<\pi(\vx)<1, \forall \vx \in  \sX$.

\textbf{\textbullet{ } 2. The presence of covariate shift due to confounding.} Even when ignorability holds because $X$ contains all confounders, a non-constant propensity score $\pi(x)$ will lead to covariate shift between the two treatment groups. When treatment effects are estimated indirectly by first obtaining estimates of $\mu_w(x)$, this can be problematic during empirical risk minimization as the observed population distribution then does not correspond to the target (marginal) distribution of characteristics.

\textbf{\textbullet{ } 3. The target label of interest is absent. } In fully randomized experiments, identifying assumptions hold by construction and the distribution of covariates across treatment arms is identical (in expectation) – yet CATE estimation remains non-trivial. This is because the true target label $Y (1) - Y (0)$ is absent even in experimental studies. Because $Y(1)$ and $Y(0)$ are available separately, outcome regressions estimating $\mu_{w(x)}=\mathbb{E}_{P}[\ry \mid \rw = w, \rvx = \vx]$ can be performed and $\hat{\tau}(x)$ can be estimated indirectly as $\hat{\mu}_1(x)$ and $\hat{\mu}_0(x)$. Nonetheless, as we discuss in Section \ref{sec:cate_estimation} it is also possible to target $\tau(x)$ directly; the two estimation strategies have different theoretical strengths \cite{curth2021nonparametric, curth2021inductive}.

\section{Feature Importance Methods} \label{appendix:feature_importance}
\subsection{Useful Properties for CATE Interpretability}

With the specificity of the CATE setting in mind, let us describe some desirable properties of the importance scores $a_i, i \in [d]$.  The presence (or absence) of these properties in popular feature importance methods is summarized in Table~\ref{tab:feature_importance_properties}.

\textbf{Sensitivity.} The covariates that do not affect the CATE model are given zero contribution. More formally, if for some $i \in [d]$ we have $\ete(\vx) = \ete(\evx_{-i})$ for all $\vx \in \sX$ , then $a_i (\ete, \vx) = 0$ for all $x \in \sX$. This allows to discard covariates that are irrelevant for the CATE model.

\textbf{Completeness.} Summing the importance scores gives the shift between the CATE and a baseline. More formally, for all $\vx \in \sX$, we have:
\begin{align*}
    \sum_{i=1}^d a_i(\ete, \vx) = \ete(\vx)  - b,
\end{align*}
where $b \in \sR$ is a constant baseline. In this way, each importance score $a_i$ can be interpreted as the contribution from covariate $i$ of $\vx$ to have a CATE that differs from the baseline $b$. Note that the choice of the baseline differs from one method to another. For instance, the baseline for SHAP and Expected Gradient~\cite{Erion2021} is the average treatment effect: $b = \Expect{\ete (\rvx)}$. Alternatively, the baseline employed in Integrated Gradient and DeepLift is the treatment effect for a baseline patient with covariates $\bar{\vx}$: $b = \ete (\bar{\vx})$. Finally, Lime uses a zero baseline: $b=0$.

\textbf{Linearity.} The importance score is linear with respect to the black-box function $f$. More formally, this means that for all covariate $i \in [d]$, real numbers $\alpha, \beta \in \sR$, functions $f_a, f_b : \sX \rightarrow \sY $ we have $a_i(\alpha \cdot f_a + \beta \cdot f_b, \cdot) = \alpha \cdot a_i(f_a, \cdot ) + \beta \cdot a_i( f_b, \cdot)$. If the CATE model $f = \ete$ is written directly in terms of the estimated potential outcomes $\ete = \epo_1 - \epo_0$, this allows to write:
\begin{align*}
    a_i(\ete, \vx) =  a_i(\epo_1, \vx) - a_i(\epo_0, \vx).
\end{align*}
This formulation renders the distinction between prognostic and predictive covariates intuitive. If $\evx_i$ is a prognostic covariate, one expects $a_i(\epo_1, \vx) = a_i(\epo_0, \vx)$ so that $a_i(\ete, \vx) = 0$, which implies that $i$ is not relevant to explain effect heterogeneity. On the other hand, if $\evx_i$ is a predictive covariate, one expects $a_i(\epo_1, \vx) \neq a_i(\epo_0, \vx)$ so that $a_i(\ete, \vx) \neq 0$, which implies that $\evx_i$ is relevant to explain effect heterogeneity.

\textbf{Model Agnosticism.} The feature importance score can be computed for all CATE model $\ete : \sX \rightarrow \sY$. Some methods only work with a restricted family of models, which prevents them from being model agnostic. A typical example is Integrated Gradient, which requires the model $\ete$ to be differentiable with respect to its input. Another example is DeepLift that requires $\ete$ to be represented by a deep-neural network.   

\textbf{Implementation Invariance.} The feature attribution would be the same for two functionally equivalent models. This means that if we have two CATE models $\ete_1$ and $\ete_2$ such that $\ete_1(x) = \ete_2(x)$ for all $x \in \sX$, this implies that $a_i(\ete_1, x) = a_i(\ete_2 , x)$ for all $x \in \sX$ and all $i \in [d]$. While this property might seem trivial, we note that it is not fulfilled when the attribution methods explicitly depend on the model's architecture. An example of such methods are the ones that use neuron activations, like DeepLift and LRP.

\begin{table*}[h]
\setstretch{1.2}
\caption{Properties of popular feature importance methods. Note that Shap also has a gradient-based implementation called GradientShap. Hence, it also belongs to the Gradient-Based category.}
\label{tab:feature_importance_properties}
\begin{center}
\begin{adjustbox}{width=\textwidth}
\begin{tabular}{ccccccc}
\toprule
\bf Type & \bf Name  & \bf Sensitivity & \bf Completeness & \bf Linearity & \bf Impl. Invariance \\
\midrule
\multirow{2}{*}{Gradient-Based} & Saliency~\cite{Simonyan2013} & \cmark & \xmark & \cmark & \cmark \\ \
 & Integrated Gradients~\cite{Sundararajan2017} & \cmark & \cmark & \cmark & \cmark \\ \cline{1-1}

\multirow{4}{*}{Perturbation-Based}&
Lime~\cite{Ribeiro2016} & \xmark & \cmark & \xmark & \cmark \\
& Feature Ablation & \cmark & \xmark & \xmark & \cmark \\
& Feature Permutation & \cmark & \xmark & \xmark & \cmark \\
& Shap~\cite{Lundberg2017} & \cmark & \cmark & \cmark & \cmark  \\ \cline{1-1}
\multirow{2}{*}{Neuron Activation} & LRP~\cite{Binder2016} & \cmark & \cmark & \cmark & \xmark \\
 & DeepLift~\cite{Shrikumar2017} & \cmark & \cmark & \cmark & \xmark \\
\bottomrule
\end{tabular}
\end{adjustbox}
\end{center}
\end{table*}

We note that two methods stand-out in the previous analysis: Integrated Gradients and Shap. The former is much more efficient computationally as it typically requires 50 backwards pass on the model per instance. Shap's complexity, on the other hand, scales exponentially with the number of features $d$~\cite{Lundberg2017}. For datasets with high $d$ (like TCGA), computing Shap for thousands of examples quickly become prohibitively expensive. For this reason, we chose Integrated Gradients as our main explanation method.

\subsection{Quantitative Comparison between Feature Importance Methods}
We shall now reproduce the experiments form Sections~\ref{sec:exp_pred_scl}~\&~\ref{sec:exp_nonlin} with different feature importance methods. To approximate Shapley values, we use the Monte-Carlo sampling from~\cite{Castro2009}. We found this approach more computationally efficient than KernelShap.

\begin{figure}
	\centering
	\begin{subfigure}[b]{.9\textwidth} 
		\centering
		\includegraphics[width=\textwidth]{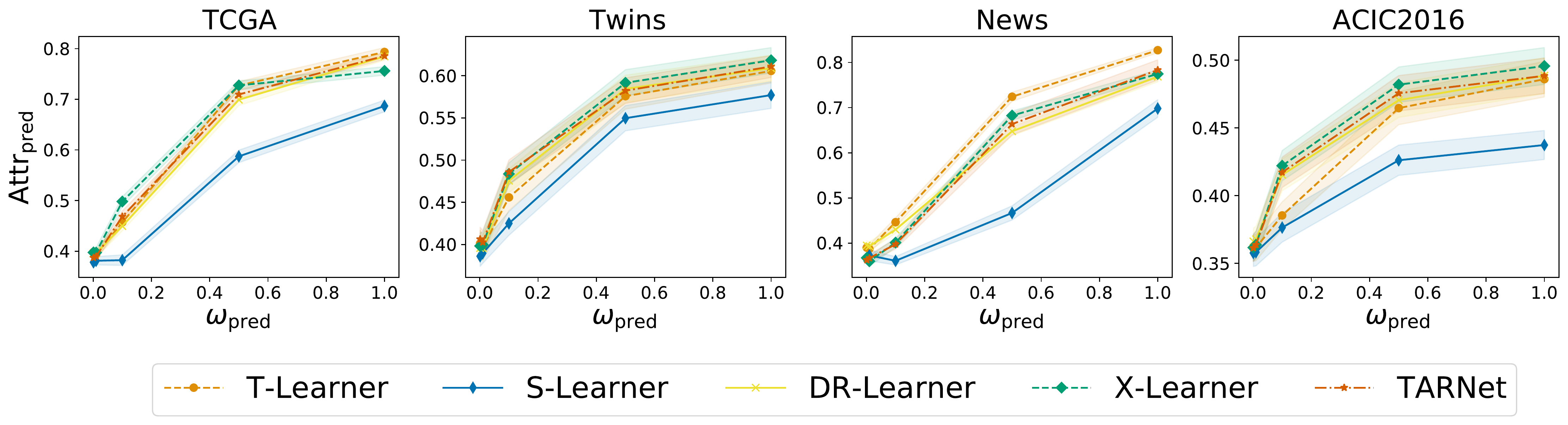}
		\caption{Integrated Gradients}
	\end{subfigure}
	\begin{subfigure}[b]{.9\textwidth} 
		\centering
		\includegraphics[width=\textwidth]{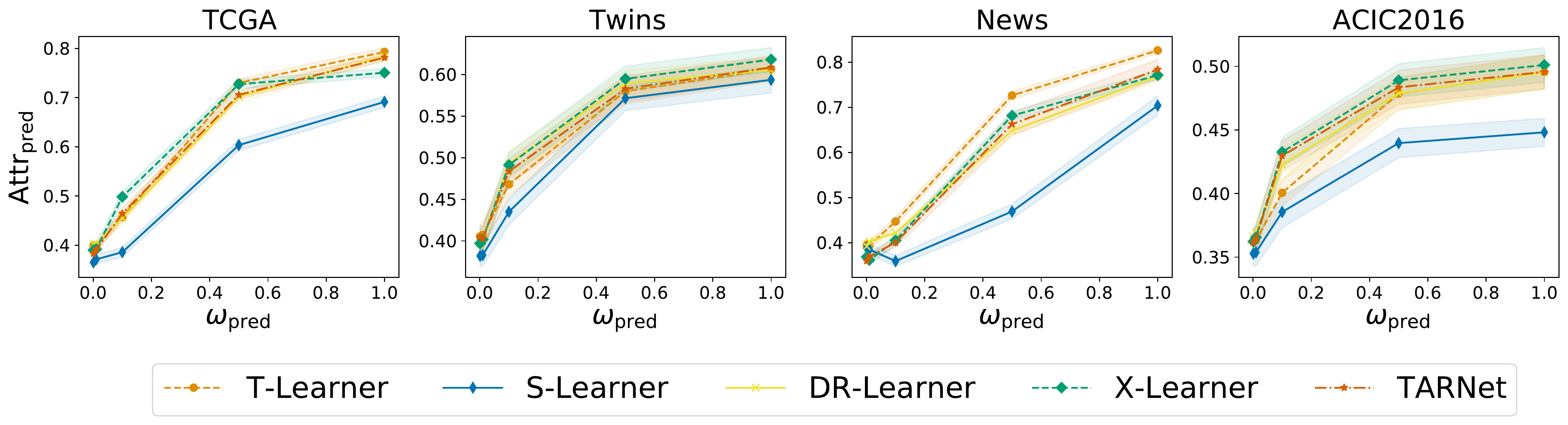}
		\caption{Shap}
	\end{subfigure}
	\begin{subfigure}[b]{.9\textwidth} 
		\centering
		\includegraphics[width=\textwidth]{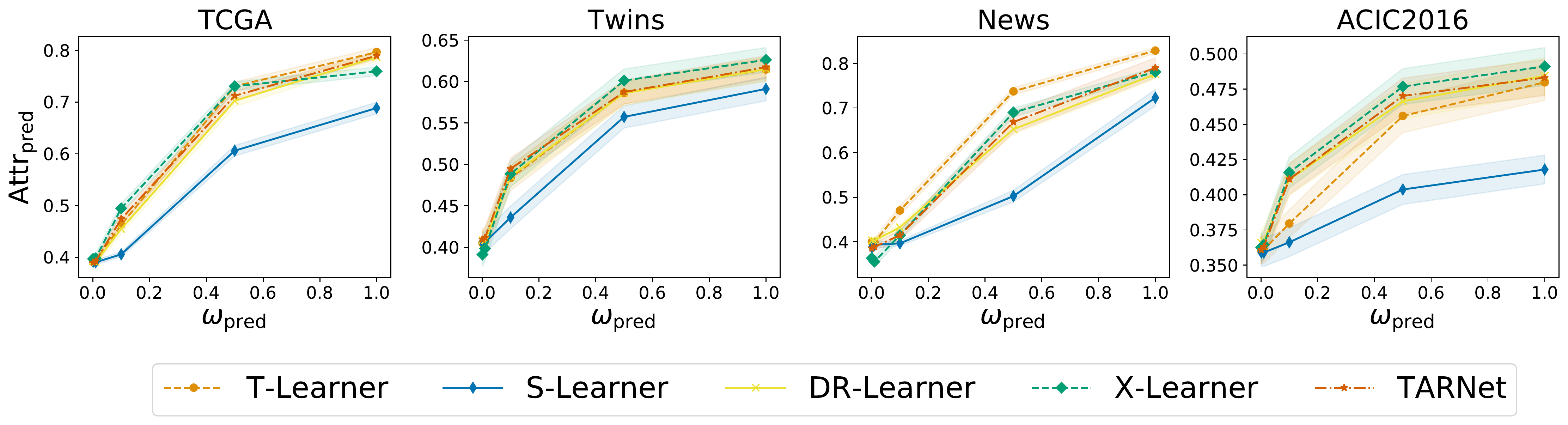}
		\caption{Feature Ablation}
	\end{subfigure}
		\begin{subfigure}[b]{.9\textwidth} 
		\centering
		\includegraphics[width=\textwidth]{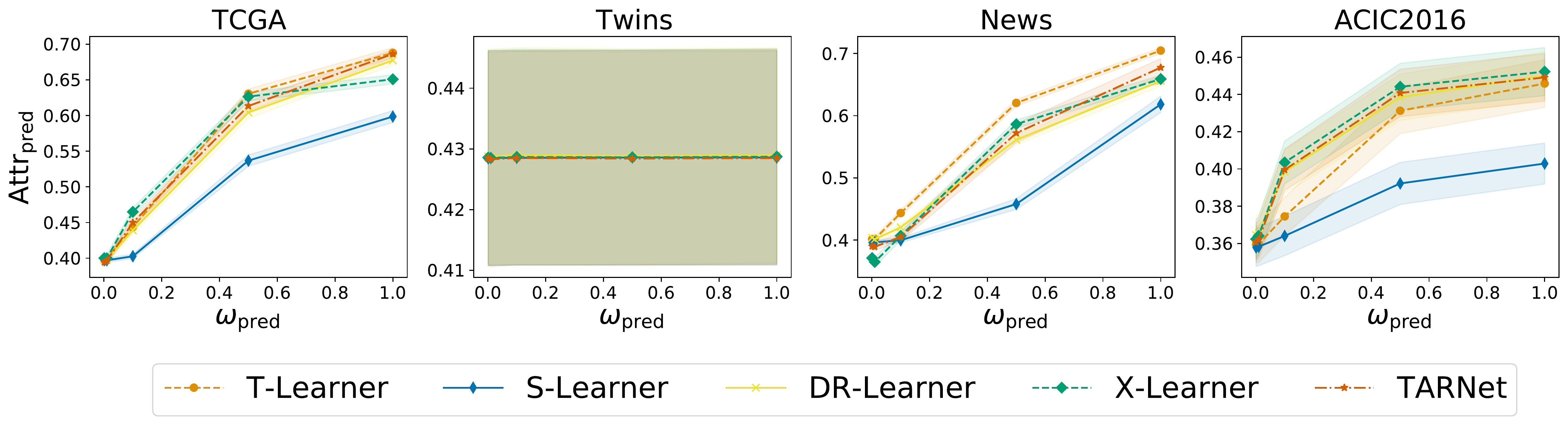}
		\caption{Feature Permutation}
	\end{subfigure}
	\caption{Performance comparison of feature importance methods in terms of $\Attrpred$ when varying the predictive scale, using four feature datasets (TGCA, Twins, News, ACIC). Averaged across multiple runs, shaded areas indicates one standard error.}
	\label{fig:predictive_scale_impact_feature_importance}
\end{figure}

\textbf{Prognostic Scale.} The experiments from Section~\ref{sec:exp_pred_scl} with various feature importance methods is reported in Figure~\ref{fig:predictive_scale_impact_feature_importance}. We clearly see that the results for Shap and Integrated Gradients are nearly identical. The predictive accuracy obtained with Feature Ablation is marginally lower, while Feature Permutation substantially underperforms. We note that all the conclusions discussed in Section~\ref{sec:exp_pred_scl} still hold if we replace Integrated Gradients by Shap or Feature Ablation. In particular, the relative ordering between learners is not affected by this choice.

\begin{figure}
	\centering
	\begin{subfigure}[b]{.9\textwidth} 
		\centering
		\includegraphics[width=\textwidth]{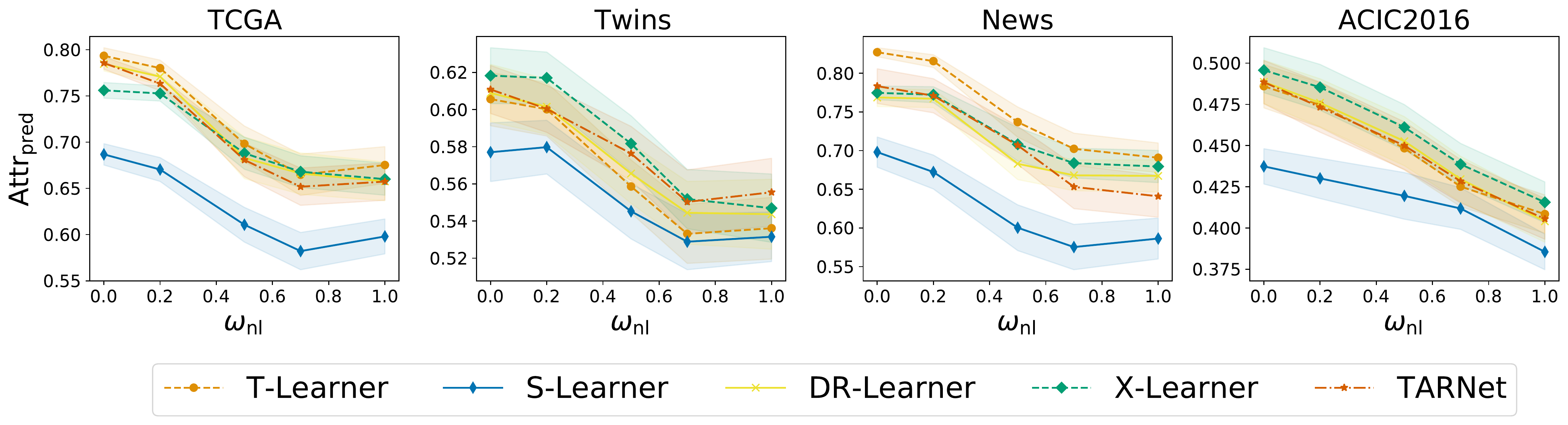}
		\caption{Integrated Gradients}
	\end{subfigure}
	\begin{subfigure}[b]{.9\textwidth} 
		\centering
		\includegraphics[width=\textwidth]{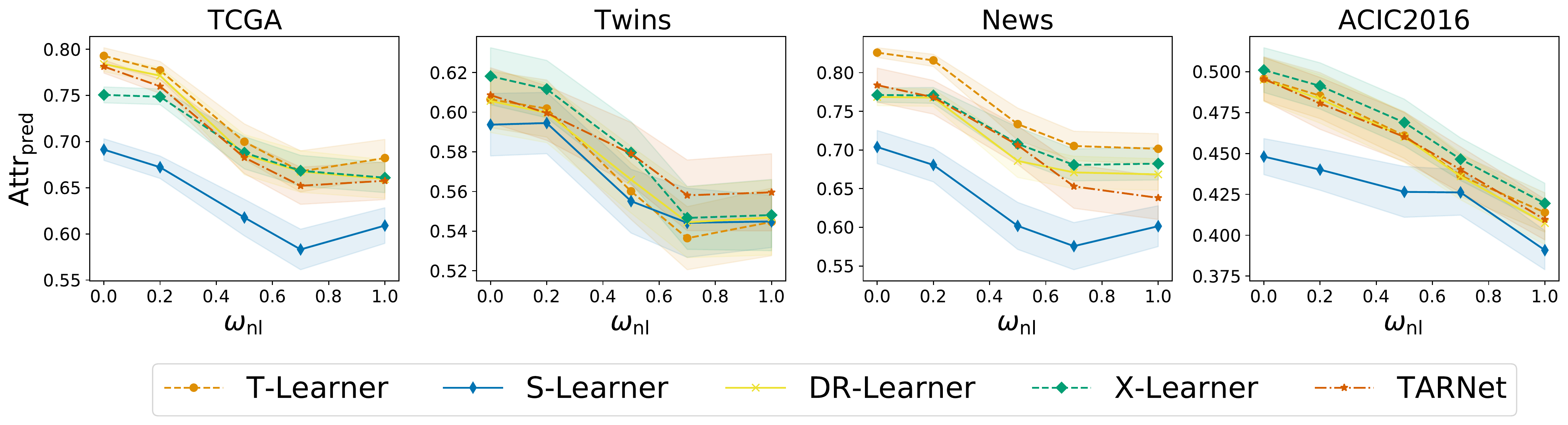}
		\caption{Shap}
	\end{subfigure}
	\begin{subfigure}[b]{.9\textwidth} 
		\centering
		\includegraphics[width=\textwidth]{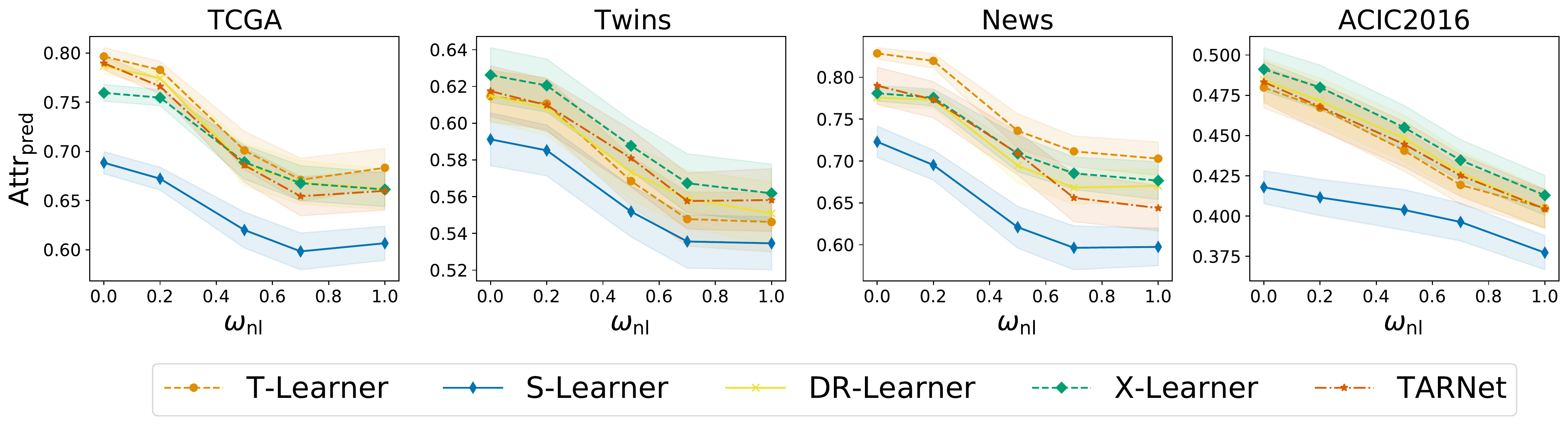}
		\caption{Feature Ablation}
	\end{subfigure}
		\begin{subfigure}[b]{.9\textwidth} 
		\centering
		\includegraphics[width=\textwidth]{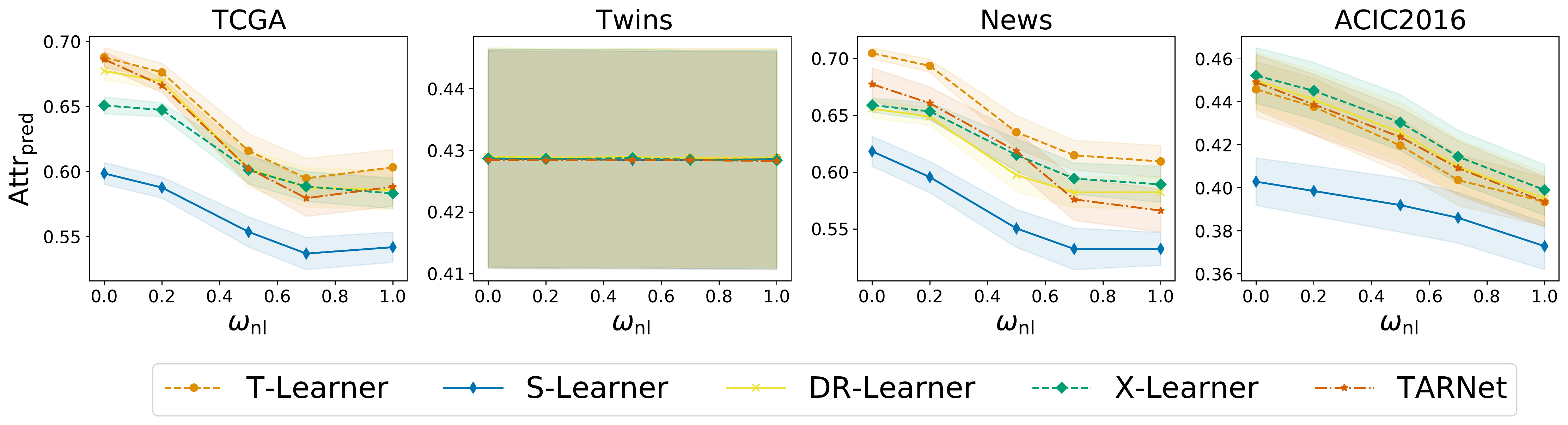}
		\caption{Feature Permutation}
	\end{subfigure}
	\caption{Performance comparison of feature importance methods in terms of $\Attrpred$ when varying the nonlinearity scale, using four feature datasets (TGCA, Twins, News, ACIC). Averaged across multiple runs, shaded areas indicates one standard error.}
	\label{fig:nonlinearity_scale_impact_feature_importance}
\end{figure}

\textbf{Nonlinearity Sensitivity.} The experiments from Section~\ref{sec:exp_nonlin} with various feature importance methods is reported in Figure~\ref{fig:nonlinearity_scale_impact_feature_importance}. Again, Shap and Integrated Gradients are closely followed by Feature Ablation and Feature Permutation significantly underperforms. All the conclusions discussed in Section~\ref{sec:exp_nonlin} still hold if we replace Integrated Gradients by Shap or Feature Ablation. In particular, the relative ordering between learners is not affected by this choice.

\section{Experimental Details and Additional Results}
\label{appendix:experiments_details}

%\hl{Official instructions for us:  For benchmarks, the supplementary materials must ensure that all results are easily reproducible. Where possible, use a reproducibility framework such as the ML reproducibility checklist, or otherwise guarantee that all results can be easily reproduced, i.e. all necessary datasets, code, and evaluation procedures must be accessible and documented.}

%\hl{Required by the paper checklist: \\
%-All the training details (e.g., data splits, hyperparameters)\\
%- Datasets: licenses \\
%- Datasets: how was consent obtained? \\
%}

%\hl{Required by us: \\
%- Extra confounding experiments
%}

\subsection{CATE Model and Implementation Details}
We use two direct learners, both of which use a first-stage regression step to estimate nuisance parameters $\eta=(\mu_0, \mu_1, \pi)$ and then use these to create a surrogate for the treatment effect in the second stage. 
In particular, \cite{Kennedy2020OptimalDR}'s DR-learner uses the pseudo-outcome
\begin{equation}
\begin{split}
\tilde{Y}_{DR, \hat{\eta}} = \left(\frac{W}{\hat{\pi}(X)}- \frac{(1-W)}{1-\hat{\pi}(X)}\right) Y + \left[\left(1 - \frac{W}{\hat{\pi}(X)}\right) \hat{\mu}_1(x)-\left(1 - \frac{1-W}{1-\hat{\pi}(X)}\right)\hat{\mu}_0(X)\right]
\end{split}
\end{equation}
which is based on the doubly-robust AIPW estimator \citep{robins1995semiparametric} and is unbiased if either propensity score \textit{or} outcome regressions are correctly specified. \cite{kunzel2019metalearners}'s X-learner, on the other hand, creates \textit{two} pseudo-outcomes, one to be used in each treatment group:  $\hat{\tau}_1$ is estimated by using pseudo outcome $Y-\hat{\mu}_0(X)$ in a regression using only treated individuals, while $\hat{\tau}_0$ is estimated by using pseudo outcome $\hat{\mu}_0(X)-Y$ in a regression using only control individuals. The two estimators are then combined using  $\hat{\tau}(x) = g(x) \hat{\tau}_1(x) + (1 - g(x)) \hat{\tau}_0(x)$ for some weighting function $g(x)$; as proposed by \cite{kunzel2019metalearners} we rely on $g(x)=\pi(x)$. 

Indirect learners (T-learner, S-learner and TARNet) are as described in the main text.

We use the PyTorch implementations of all models provided in the python package \texttt{Catenets}\footnote{\url{https://github.com/AliciaCurth/CATENets}}; as \cite{curth2021nonparametric} we ensure that each estimated function ($\hat{\mu}_w(x)$, $\hat{\pi}(x)$ and $\hat{\tau}(x)$) has access to the same amount of hidden layers and units in total (2 hidden layers with 100 units and a final prediction layer; for TARNet/CFRNet this means that the representation $\Phi$ and the outcome heads $h_w$ each have 1 hidden layer)  and each architecture can hence represent similarly complex nuisance functions. We use dense layers with ReLU activation function. All models are trained using the Adam optimizer with learning rate $10^{-4}$, batch size 1024 and early stopping on the validation set (which represents $30\%$ of the initial training set). 

We used a virtual machine with 6 CPUs, an Nvidia K80 Tesla GPU and 56GB of RAM to run all experiments. 

\subsection{Dataset Details}

\textbf{TCGA.}  The TCGA dataset \cite{weinstein2013cancer} consists of information about gene expression measurements from $9659$ cancer patients. We use the same version of the TCGA dataset as in \cite{schwab2019learning}\footnote{\url{https://github.com/d909b/drnet}}. In our experiments, we use as patient covariates the measurements from the $100$ most variable genes. These are all continuous features. The data is log-normalized and each feature is scaled in the $[0, 1]$ interval. 

\textbf{Twins.} The Twins dataset \cite{almond2005costs} consists of information from $11400$ twin births in the USA recorded between 1989-1991. Each twin pair is characterized by $39$ covariates related to the parents, pregnancy and birth; these represent a mixture of continuous and categorical features. In our experiments, the publicly available version of the dataset from \texttt{Catenets} is used where we randomly sample one of the twins to observe. 

\textbf{News.} The News dataset consists of 10000 news items (randomly sampled), each characterized by 2858 word counts \cite{newman2008bag, johansson2016learning, schwab2019learning}. Similarly to \cite{johansson2016learning} we perform Principal Component Analysis and use as covariates for each news item the first 100 principal components (continuous features). We use the same version of the News dataset as used in \cite{schwab2019learning}.

\textbf{ACIC2016.} The ACIC2016 dataset consists of data from the Collaborative Perinatal Project provided as part of the Atlantic Causal
Inference Competition (ACIC2016) \cite{dorie2019automated}. We use the publicly available version of the dataset from the \texttt{Catenets} package which consist of 55 covariates (mixture of continuous and categorical ones) for 2200 patients. Note that the same version of the dataset was used in \cite{curth2021inductive}. 

For information about how each dataset was collected and curated, refer to the corresponding references. Note that all of these datasets are publicly available. Each dataset undergoes a 80\%/20\% split for training/testing respectively. Moreover 30\% of the training dataset is used for validation as part of the early stopping procedure performed by the \texttt{Catenets} package. The feature importance metrics are computed for up to 1000 examples from the test set, while the PEHE is computed over the entire test set. The results are averaged over 30 random seeds for Experiments 1 \& 2 and over 10 random seeds for Experiment 3 (note that there are three experimental settings here for the different types of confounding). 

\subsection{Feature Importance Methods Implementation}
We use the Pytorch implementation of all feature importance methods provided in the Python package \texttt{Captum}\footnote{\url{https://captum.ai/}}. For the relevant feature importance methods, we set the zero vector as a baseline input: $\bar{x} = 0$.

%\subsection{Experiment 1: Altering the Strength of Predictive Effects}
%\hl{anything to add here?}

\subsection{Experiment 2: Incorporating Nonlinearities}

Each nonlinear function $\chi$ is sampled randomly from the function set $\mathcal{F} = \{ \chi_1 : x \mapsto \abs{x}, \ \chi_2: x \mapsto \exp(-x^2), \ \chi_3 : x \mapsto (1 + x^2)^{-1}, \ \chi_4: x \mapsto  \cos(x), \ \chi_5: x \mapsto \sin(x) , \ \chi_6: x \mapsto  \arctan(x) , \ \chi_7: x \mapsto \tanh(x), \ \chi_8: x \mapsto \log(1+x^2), \ \chi_9: x \mapsto (1+x^2)^{\nicefrac{1}{2}}, \ \chi_{10}: x \mapsto \cosh(x) \}$.

\subsection{Experiment 3: The Effect of Confounding}
\begin{figure}
	\centering
	\begin{subfigure}[b]{.6\textwidth} 
		\centering
		\includegraphics[width=\textwidth]{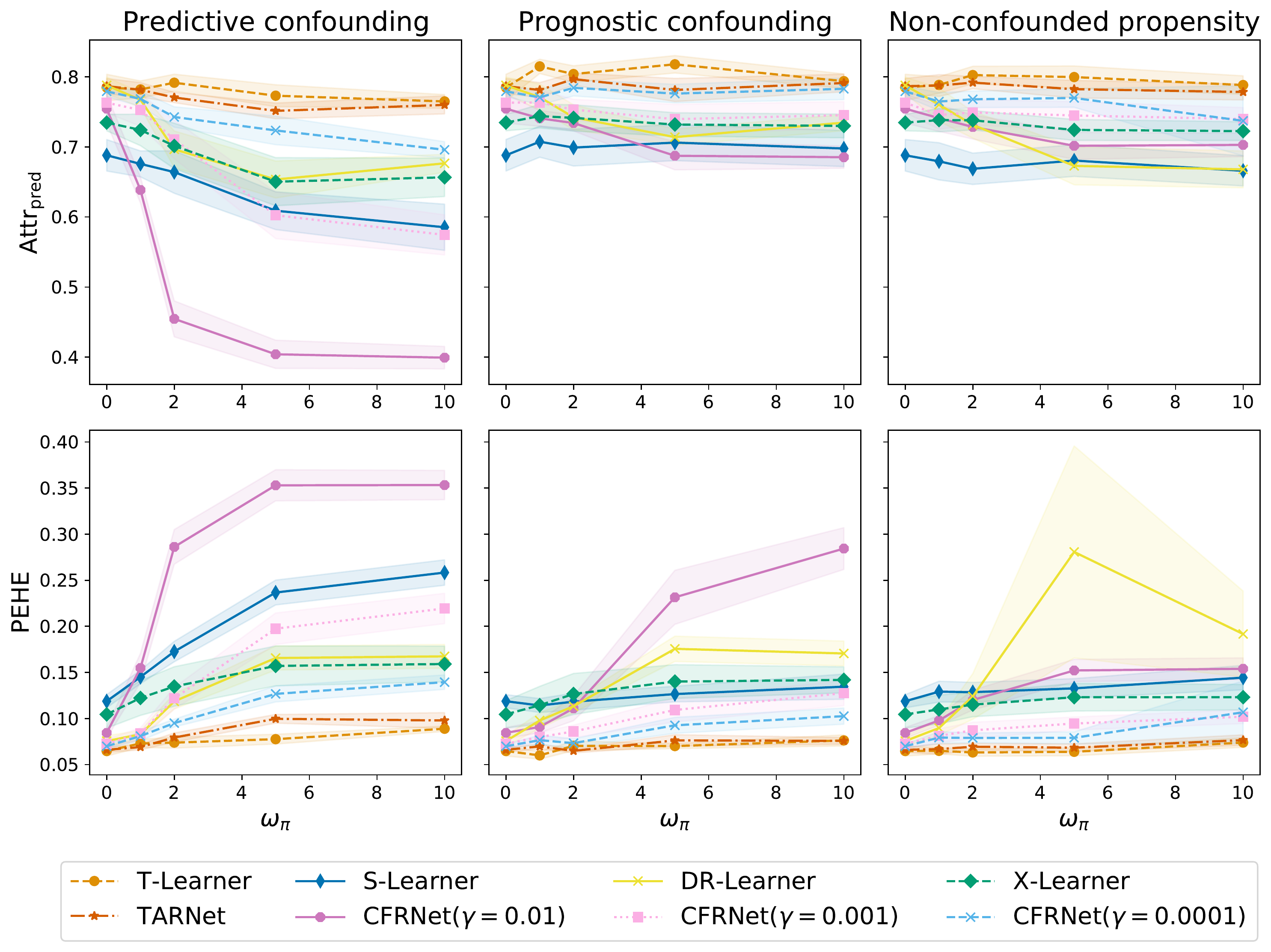}
		\caption{TCGA}
	\end{subfigure}
	\begin{subfigure}[b]{.6\textwidth} 
		\centering
		\includegraphics[width=\textwidth]{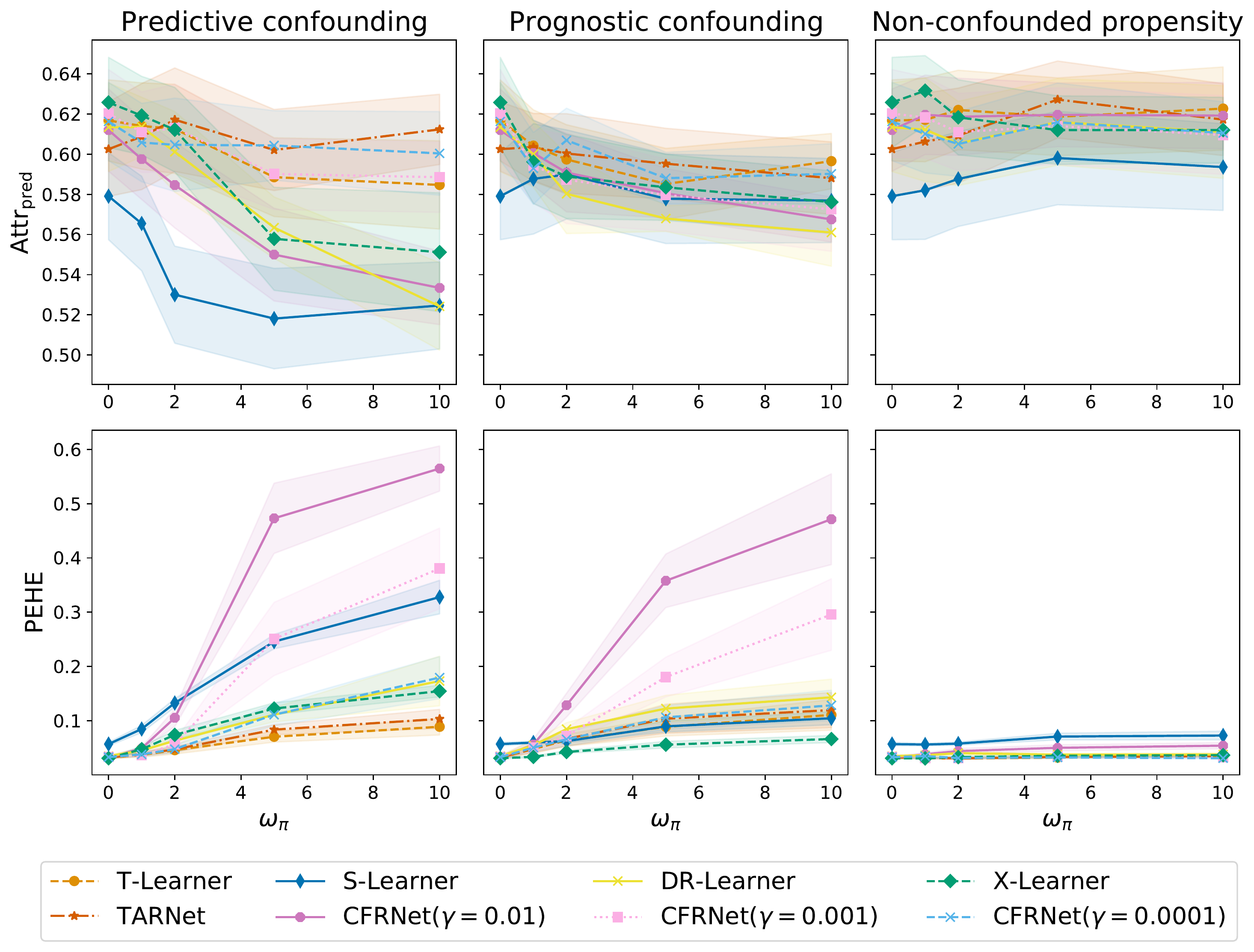}
		\caption{Twins}
	\end{subfigure}
	\begin{subfigure}[b]{.6\textwidth} 
		\centering
		\includegraphics[width=\textwidth]{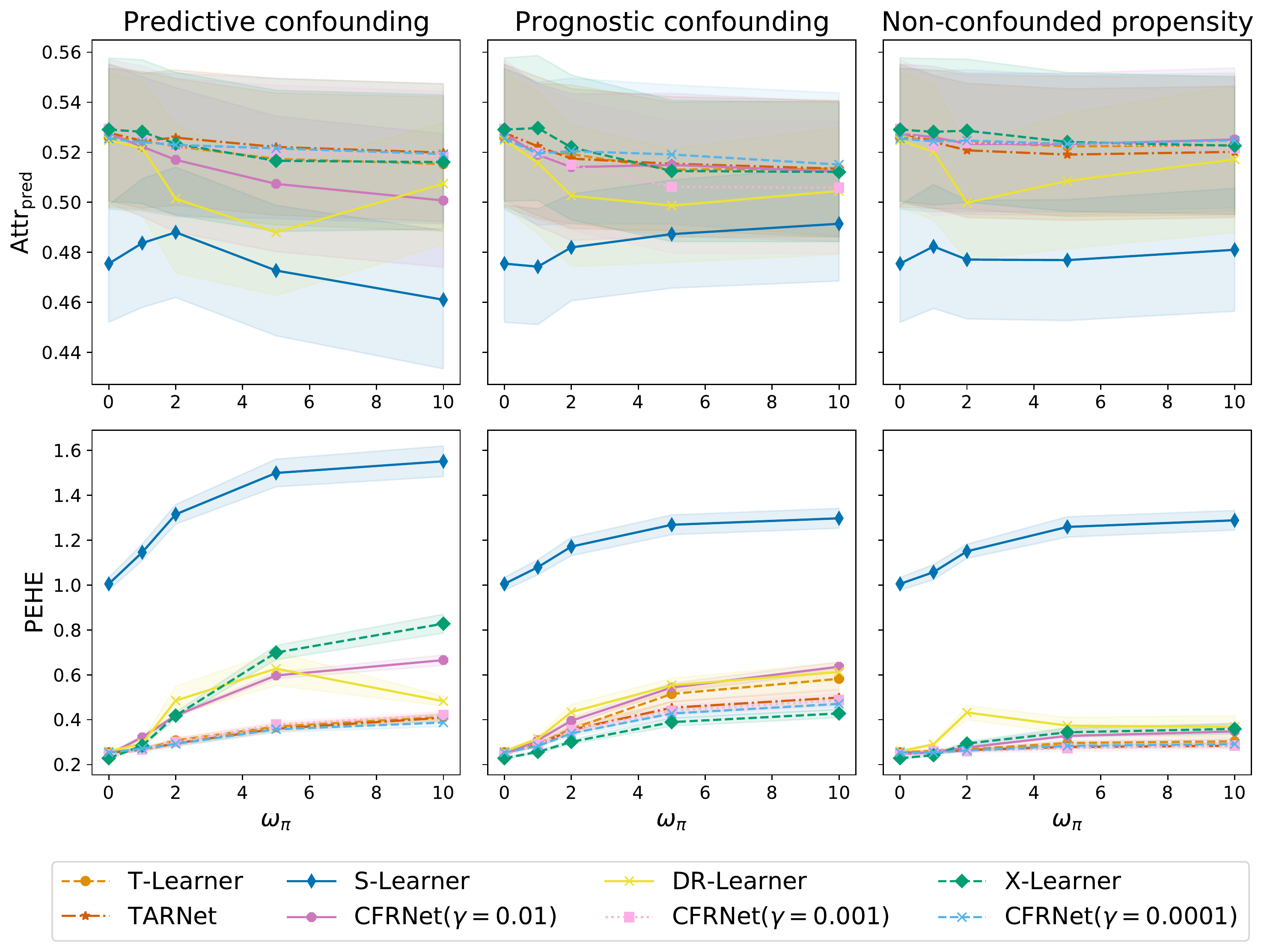}
		\caption{ACIC2016}
	\end{subfigure}
	\caption{Performance comparison when increasing the propensity scale. Averaged across multiple runs, shaded areas indicates one standard error.}
	\label{fig:propensity_scale_all_datasets}
\end{figure}

In Fig. \ref{fig:propensity_scale_all_datasets} we present results for the effect of confounding for the three datasets not covered in the main text. Many observations we originally made using the News dataset carry over to the other datasets; in particular, we observe that \circled{1} attribution trends in the predictive confounding setting vary with $\omega_\pi$, \circled{2} there remain systematic differences across the three confounding settings (attributions worsen mainly in the predictive confounding setting, PEHE worsens in predictive and prognostic confounding settings) and \circled{3} balancing regularization can \textit{harm} performance when confounding is predictive or prognostic. The main difference when compared to the results in the main text is that across all other datasets we observe less deterioration in attribution accuracy in the non-confounded propensity setting, and therefore see less of a difference between prognostic confounding and non-confounded propensity when considering only  $\Attrpred$ (this changes when considering also PEHE).

\end{document}